\documentclass[journal]{IEEEtran}

\usepackage{cite}
\ifCLASSINFOpdf
  \usepackage[pdftex]{graphicx}
  \DeclareGraphicsExtensions{.pdf,.jpeg,.png}
\else
  \usepackage[dvips]{graphicx}
  \DeclareGraphicsExtensions{.eps}
\fi
\usepackage[cmex10]{amsmath}
\usepackage{amssymb,amsfonts,amsthm,mathtools}
\usepackage{algorithmic,algorithm}
\usepackage{array,multirow}
\ifCLASSOPTIONcompsoc
  \usepackage[caption=false,font=normalsize,labelfont=sf,textfont=sf]{subfig}
\else
  \usepackage[caption=false,font=footnotesize]{subfig}
\fi
\usepackage{url}

\DeclareMathOperator*{\argmin}{arg\,min}
\DeclareMathOperator*{\argmax}{arg\,max}

\begin{document}

\title{Adaptive-Rate Compressive Sensing Using Side Information}


\author{Garrett~Warnell,
		Sourabh~Bhattacharya,
		Rama~Chellappa,
		and~Tamer~Ba\c{s}ar
\thanks{G. Warnell and R. Chellappa are with the University of Maryland College Park, College Park, MD}%
\thanks{S. Bhattacharya is with Iowa State University, Ames, IA}%
\thanks{T. Ba\c{s}ar is with the University of Illinois Urbana-Champaign, Urbana, IL}}%

\maketitle

\begin{abstract}
We provide two novel adaptive-rate compressive sensing (CS) strategies for sparse, time-varying signals using side information.  Our first method utilizes extra \emph{cross-validation} measurements, and the second one exploits extra \emph{low-resolution} measurements.  Unlike the majority of current CS techniques, we do not assume that we know an upper bound on the number of significant coefficients that comprise the images in the video sequence.  Instead, we use the side information to predict the number of significant coefficients in the signal at the next time instant.  For each image in the video sequence, our techniques specify a fixed number of spatially-multiplexed CS measurements to acquire, and adjust this quantity from image to image.  Our strategies are developed in the specific context of background subtraction for surveillance video, and we experimentally validate the proposed methods on real video sequences.
\end{abstract}

\IEEEpeerreviewmaketitle

\section{Introduction}
\label{secn::intro}
Visual surveillance is a task that often involves collecting a large amount of data in search of information contained in relatively small segments of video.  For example, a surveillance system tasked with intruder detection will often spend most of its time collecting observations of a scene in which no intruders are present.  Without any such \emph{foreground objects}, the corresponding surveillance video is useless: it is only the portions of video that depict these unexpected objects in the environment that are useful for surveillance.  However, because it is unknown when such objects will appear, many systems gather the same amount of data regardless of scene content.  This static approach to sensing is wasteful in that resources are spent collecting unimportant data.  However, it is not immediately clear how to efficiently acquire useful data since the periods of scene activity are unknown in advance.  If this information were available \emph{a priori}, a better scheme would be to collect data only during times when foreground objects are present.

In any attempt to do so, the system must make some sort of real-time decision regarding scene activity.  However, such a decision can be made only if real-time data to that effect is available.  We shall refer to such data as \emph{side information}.  Broadly, this information can come from two sources: a secondary modality and/or the primary video sensor itself.  In this paper, we develop two adaptive sensing schemes that exploit side information that comes from an example of each.  Our first strategy employs a single video sensor to continuously make observations that are simultaneously used to infer both the foreground and the scene activity.  The second adaptive method we present determines scene activity using observations that come from a secondary visual sensor.  Both methods utilize a \emph{compressive sensing} (CS) \cite{Candes2006} \cite{Donoho2006} \cite{Candes2006a} \cite{Candes2006b} \cite{Baraniuk2007} camera as the primary modality.  While many such sensors are beginning to emerge \cite{Willett2011}, our methods are specifically developed for a fast variant of a spatially multiplexing camera such as the single-pixel camera \cite{Duarte2008} \cite{Romberg2008}.

\begin{figure*}
        \centering
        \subfloat[]{\includegraphics[width=0.2\textwidth]{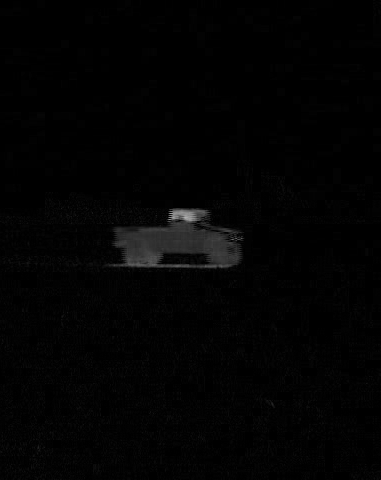} \label{fig::intro_measurementrate_true}}
        ~
        \subfloat[]{\includegraphics[width=0.2\textwidth]{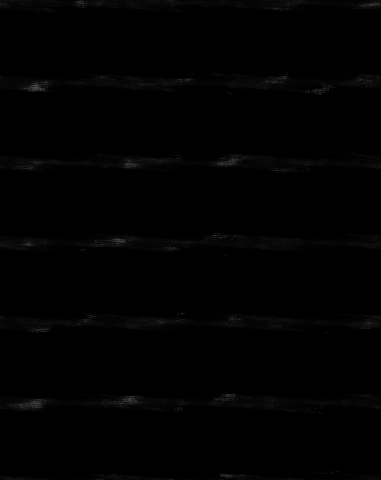} \label{fig::intro_measurementrate_low}}
        ~
        \subfloat[]{\includegraphics[width=0.2\textwidth]{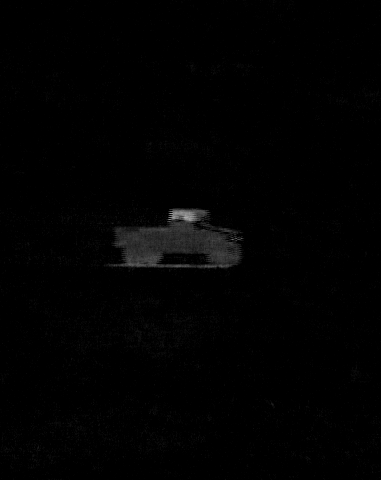} \label{fig::intro_measurementrate_opt}}
        ~
        \subfloat[]{\includegraphics[width=0.2\textwidth]{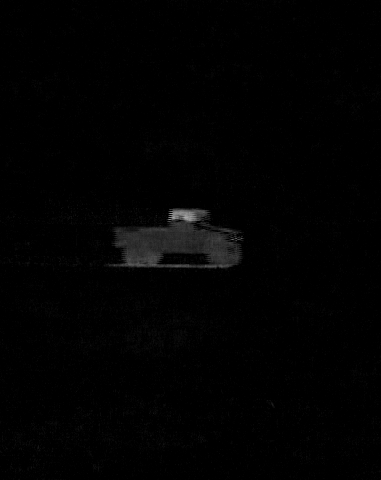} \label{fig::intro_measurementrate_high}}
        \caption{\footnotesize{Foreground reconstruction with varying measurement rates. \protect\subref{fig::intro_measurementrate_true} is the true foreground, \protect\subref{fig::intro_measurementrate_low} is the foreground reconstruction when too few measurements are used, \protect\subref{fig::intro_measurementrate_opt} is the reconstruction when an optimal number of measurements are used, and \protect\subref{fig::intro_measurementrate_high} is the reconstruction when more than the optimal number of measurements are used.}}
        \label{fig::intro_measurementrate}
\end{figure*}

In this paper, we consider the following basic scenario: a CS camera is tasked with observing a region for the purpose of obtaining foreground video.  Since the foreground often occupies only a relatively small number of pixels, Cevher \emph{et al.} \cite{Cevher2008} have shown that a small number of compressive measurements provided by this camera are sufficient to ensure that the foreground can be accurately inferred.  However, the solution provided in that work implicitly relies on an assumption that is pervasive in the CS literature: that an upper bound on the \emph{sparsity} (number of significant components) of the signal(s) under observation is known.  Such an assumption enables the use of a static measurement process for each image in the video sequence.  However, foreground video is a dynamic entity: changes in the number and appearance of foreground objects can cause large changes in sparsity with respect to time.  Underestimating this quantity will lead to the use of a CS system that will provide too few measurements for an accurate reconstruction.  Overestimating signal sparsity, on the other hand, will require the collection of more measurements than necessary to achieve such a reconstruction.  For example, consider Figure \ref{fig::intro_measurementrate}.  The true foreground's (Figure \ref{fig::intro_measurementrate}\subref{fig::intro_measurementrate_true}) reconstruction is poor when too few compressive measurements are collected (Figure \ref{fig::intro_measurementrate}\subref{fig::intro_measurementrate_low}), but looks virtually the same whether or not an optimal or greater-than-optimal number of measurements are acquired (Figures \ref{fig::intro_measurementrate}\subref{fig::intro_measurementrate_opt} and \ref{fig::intro_measurementrate}\subref{fig::intro_measurementrate_high}, respectively).  Therefore, dependent on the number of measurements acquired at each time instant, the static CS approach is insufficient at worst and wasteful at best.

We provide in this paper novel, adaptive-rate CS strategies that seek to address this problem.  The approaches we present utilize two different forms of side information: \emph{cross-validation measurements} and \emph{low-resolution measurements}.  In each case, we use the extra information in order to predict the number of foreground pixels (sparsity) in the next frame.

\subsection{Related Work}
\label{secn::intro_relatedwork}
Adapting the standard CS framework to a dynamic, time-varying signal is something that has been studied from various perspectives by several researchers.

Wakin \emph{et al.} \cite{Wakin2006}, Park and Wakin \cite{Park2009}, Sankaranarayanan \emph{et al.} \cite{Sankaranarayanan2012}, and Reddy \emph{et al.} \cite{Reddy2011} have each proposed video-specific versions of CS.  Each one leverages video-specific signal dynamics such as temporal correlation and optical flow.  For measurement models that provide streaming CS measurements, Sankaranarayan \emph{et al.} \cite{Sankaranarayanan2013a}, Asif and Romberg \cite{Asif2013}, and Angelosante \emph{et al.} \cite{Angelosante2010} have proposed adaptive CS decoding procedures that are faster and more accurate than those that do not explicitly model the video dynamics.

Vaswani \emph{et al.} \cite{Vaswani2008} \cite{Vaswani2010} \cite{Vaswani2010a}, Cossalter \emph{et al.} \cite{Cossalter2010}, and Stankovic \emph{et al.} \cite{Stankovic2009} \cite{Stankovic2009a} propose modifications to the CS decoding step that leverage extra signal support information in order to provide more accurate reconstructions from a fixed number of measurements. More generally, Scarlett \emph{et al.} \cite{Scarlett2013} provide generic information-theoretic bounds for any support-adaptive decoding procedure.  Malioutov \emph{et al.} \cite{Malioutov2010} and Boufonous \emph{et al.} \cite{Boufounos2007} propose decoders with adaptive stopping criteria: sequential signal estimates are made until either a consistency or cross-validation criterion is met.

Several researchers have also considered adaptive encoding techniques.  These techniques primarily focus on finding and using the ``best" compressive measurement vectors at each instant of time.  Ashok \emph{et al.} \cite{Ashok2008} propose an offline procedure in order to design entire measurement matrices optimized for a specific task.  Similarly, Duarte-Carvajalino \emph{et al.} \cite{Duarte-Carvajalino2012} compute class-specific optimal measurements offline, but decide which class to use using an online procedure with a fixed number of measurements.  Purely online procedures include those developed by Averbuch \emph{et al.} \cite{Averbuch2012}, Ji \emph{et al.} \cite{Ji2008}, Chou \emph{et al.} \cite{Chou2009}, and Haupt \emph{et al.} \cite{Haupt2012}: the next-best measurement vectors are computed by optimizing criterion functions that seek to minimize quantites such as posterior entropy and expected reconstruction error.  Some of these methods use a fixed measurement rate, while others propose stopping criterion similar to several of the adaptive decoding procedures.

Some of the above methods exhibit an adaptive measurement rate in that they stop collecting measurements when certain criteria are met.  However, due to the dynamic nature of video signals, it may not be possible to evaluate these criteria (as they often involve CS decoding) and collect a new measurement before the signal has significantly changed.  Recent adaptive-rate work by Yuan \emph{et al.} \cite{Yuan2013} and Schaeffer \emph{et al.} \cite{Schaeffer} sidesteps this problem by using a static spatial measurement rate and considering how to adaptively select the \emph{temporal} compression rate through batch analysis.  In contrast, we propose here techniques that specify a fixed number of \emph{spatially}-multiplexed measurements to acquire before sensing the signal at a given time instant and modify this quantity between each acquisition without assuming that the signal remains static between acquisitions.  That is, we consider a system in which the decoding procedure is fixed and we are able to change the encoding procedure, which is fundamentally different from the previously-discussed work on adaptive decoding procedures (e.g., that of Vaswani \emph{et al.} \cite{Vaswani2008} \cite{Vaswani2010} \cite{Vaswani2010a}).

\subsection{Organization}
\label{secn::intro_organization}
This paper is organized as follows.  In Section \ref{secn::cs}, we provide a brief overview of CS.  Sections \ref{secn::problemstatement} and \ref{secn::backgroundsubtraction} contain a precise formulation of and context for our rate-adaptive CS algorithms.  Our measurement acquisition technique is described in Section \ref{secn::sensingmatrixdesign}.  The proposed adaptive rate CS techniques are discussed in Sections \ref{secn::method1} and \ref{secn::method2}, and they are experimentally validated in Section \ref{secn::experiments}.  Finally, we provide a summary and future research directions in Section \ref{secn::summary}.
\section{Compressive Sensing}
\label{secn::cs}
Compressive sensing is a relatively new theory in sensing which asserts that a certain class of discrete signals can be adequately sensed by capturing far fewer measurements than the dimension of the ambient space in which they reside.  By ``adequately sensed," it is meant that the signal of interest can be accurately inferred using the measurements acquired by the sensor.

In this paper, we use CS in the context of imaging.  Consider a grayscale image $F \in \mathbb{R}^{N \times N}$, vectorized in column-major order as $\mathbf{f} \in \mathbb{R}^{N^2}$.  A traditional camera uses an $N \times N$ array of photodetectors in order to produce $N^2$ measurements of $F$: each detector records a single value that defines the corresponding component of $\mathbf{f}$.  If we are instead able to gather measurements of a fundamentally different type, CS theory suggests that we may be able to determine $\mathbf{f}$ from far fewer than $N^2$ of them.  Specifically, these \emph{compressive measurements} record linear combinations of pixel values, i.e., $\boldsymbol{\xi} = \mathbf{\Phi}\mathbf{f}$, where $\mathbf{\Phi} \in \mathbb{C}^{M \times N^2}$ is referred to as a \emph{measurement matrix} and $M<<N^2$.

CS theory presents three general conditions under which the above claim is valid.  First, $\mathbf{f}$ should be \emph{sparse} or \emph{compressible}.  In general, a vector is said to be sparse if very few of its components are nonzero; more precisely, vectors having no more than $s$ nonzero components are said to be \emph{$s$-sparse}.  A vector is said to be compressible if it is well-approximated by a sparse signal, i.e., it has a small number of components with a large magnitude and many with much smaller magnitudes.

Second, the measurement matrix (encoder) should exhibit the \emph{restricted isometry property (RIP)} of a certain order and constant.  Specifically, $\mathbf{\Phi}$ exhibits the RIP of order $s$ with constant $\delta_s$ if the following inequality holds for all $s$-sparse $\mathbf{f}$: 
\begin{align}
(1-\delta_s) \leq \frac{\lVert \mathbf{\Phi}\mathbf{f} \rVert_2^2}{\lVert \mathbf{f} \rVert_2^2} \leq (1+\delta_s) \quad .
\end{align}
While we will discuss proposed construction methods for a $\mathbf{\Phi}$ that exhibits the RIP for specified $s$ and $\delta_s$ in Section \ref{secn::sensingmatrixdesign}, they generally involve selecting $M$ such that it exceeds a lower bound that grows with increasing $s$ and decreasing $\delta_s$.

Finally, an appropriate decoding procedure, $\hat{\mathbf{f}} = \Delta(\boldsymbol{\xi},\mathbf{\Phi})$, should be used.  While many successful decoding schemes have been discussed in the literature, we shall focus here on one in particular:
\begin{align}
\Delta(\boldsymbol{\xi},\mathbf{\Phi}) = \argmin_{\mathbf{z}\in\mathbb{R}^{N^2}} \; \lVert \mathbf{z} \rVert_1 \; \text{subject to} \; \mathbf{\Phi}\mathbf{z} = \boldsymbol{\xi} \quad ,
\label{eqn::cs_l1decoding}
\end{align}
where the $\ell_1$ norm is given explicitly by $\lVert \mathbf{z} \rVert_1 = \sum_i \vert z(i) \vert $.

With these three conditions in mind, CS theory provides us with the following result: for an $s$-sparse $\mathbf{f}$ measured with a $\mathbf{\Phi}$ that exhibits the RIP of order $2s$ with $\delta_{2s} \leq \sqrt{2}-1$, $\Delta(\boldsymbol{\xi},\mathbf{\Phi})$ will exactly recover $\mathbf{f}$ \cite{Baraniuk2011a}.  If $\mathbf{f}$ is compressible, a similar result that bounds the reconstruction error is available.  Thus, by modifying the sensor and decoder to implement $\mathbf{\Phi}$ and $\Delta$, respectively, $\mathbf{f}$ can be adequately sensed using only $M<<N^2$ measurements.

Sensors based on the above theory are still just beginning to emerge \cite{Willett2011}.  One of the most notable is the single-pixel camera \cite{Duarte2008}, where measurements specified by each row of $\mathbf{\Phi}$ are sequentially computed in the optical domain via a digital micromirror device and a single photodiode.  Throughout the remainder of this paper, we shall assume that such a device is the primary sensor.
\section{Problem Statement}
\label{secn::problemstatement}
We assume that we possess a CS camera that is capable of acquiring a variable number of compressive measurements at discrete instants of time.  We denote the measurement matrix at time $t$ by $\mathbf{\Phi}_t \in \mathbb{R}^{M_t \times N^2}$, and we construct it via  a process that depends only on our choice for $M_t$ (see Section \ref{secn::sensingmatrixdesign}).  The value used for $M_t$ will be determined by the adaptive sensing strategy prior to time $t$.  The images we observe will be of size $N \times N$, and $X_t \in \mathbb{R}^{N \times N}$ will denote the specific image at time $t$.  Vectorizing $X_t$ using column-major order as $\mathbf{x}_t \in \mathbb{R}^{N^2}$ allows us to write the compressive measurement process at time $t$ as $\mathbf{y}_t = \mathbf{\Phi}_t \mathbf{x}_t$.

We will present two adaptive sensing strategies that will each exploit a different type of side information.  The first strategy uses a small set of \emph{cross-validation measurements}, $\boldsymbol{\chi}_t \in \mathbb{R}^r$ obtained from a static linear measurement operator $\mathbf{\Psi} \in \mathbb{C}^{r \times N^2}$, i.e., $\boldsymbol{\chi}_t = \mathbf{\Psi} \mathbf{x}_t$.  $\mathbf{\Psi}$ here is referred to as a \emph{cross-validation matrix}.  The second strategy we present relies on a set of \emph{low-resolution measurements}, $Z_t \in \mathbf{R}^{L \times L}$ that we obtain via a secondary sensor that collects lower-resolution measurements of $X_t$.  Such multi-camera systems are not uncommon in the surveillance literature (see, e.g., \cite{Clady2001} \cite{Senior 2005}).

Having established the above notation, the problem we address in this paper is that of how to use the observations $\mathbf{y}_t$, $\boldsymbol{\chi}_t$, and $Z_t$ to select a minimal value for $M_{t+1}$ that will ensure $\mathbf{\Phi}_{t+1}$ gathers enough information to ensure accurate reconstruction of the foreground (dynamic) component of the high-resolution $X_t$.
\section{Compressive Sensing for Background Subtraction}
\label{secn::backgroundsubtraction}
We present our work in the context of the problem of background subtraction for video sequences.  Broadly, background subtraction is the process of decomposing an image into foreground and background components, where the foreground usually represents the objects of interest in the environment under observation.  For our purposes, we shall adopt the following model for images $\mathbf{x}_t$:
\begin{align}
\mathbf{x}_t = \mathbf{f}_t + \mathbf{b} \quad ,
\label{eqn::backgroundsubtraction_signalmodel}
\end{align}
where $\mathbf{b}$ is an unknown but deterministic static component of each image in the video sequence and $\mathbf{f}_t$ is a random variable.  At time $t$, we estimate the locations of foreground pixels by computing the set of indices $\mathcal{F}_t = \left\{ i \; : \; \vert f_t(i) \vert \geq \tau \right\}$, for some pre-defined threshold $\tau$.  We further assume that the components of $\mathbf{f}_t$ that correspond to $\mathcal{F}_t$ are bounded in magnitude, i.e.,  $\vert f_t(i) \vert \leq 1$ for all $i \in \mathcal{F}_t$.

Throughout this work, we shall assume that the components of $\mathbf{f}_t$ are distributed as follows:
\begin{align}
f_t(i) \sim \begin{cases} \mathbf{U}\left\{ \left[-1,-\tau \right] \cup \left[ \tau,1 \right] \right\} & , i \in \mathcal{F}_t \\
\mathcal{N}(0,\sigma_b^2) & , i \not\in \mathcal{F}_t \end{cases} \quad ,
\label{eqn::backgroundsubtraction_foregroundmodel}
\end{align}
where each component is assumed to be independent of the others.  We have approximated the intensity distribution of those pixels not in $\mathcal{F}_t$ as a zero-mean Gaussian under the assumption that $\sigma_b^2$ is much smaller than $\tau$.

Following the work of Cevher \emph{et al.} \cite{Cevher2008}, we seek to perform background subtraction in the compressive domain.  Often, it is the case that the foreground occupies only a very small portion of the image plane, i.e., $\vert \mathcal{F}_t \vert << N^2$.  Given the foreground model (\ref{eqn::backgroundsubtraction_foregroundmodel}), this implies that $\mathbf{f}_t$ is compressible in the spatial domain.  Therefore, if $\mathbf{b}$ is known, we can use it, (\ref{eqn::backgroundsubtraction_signalmodel}), and compressive image measurements $\mathbf{y}_t = \mathbf{\Phi}_t\mathbf{x}_t$ to generate the following estimate of $\mathbf{f}_t$:
\begin{align}
\hat{\mathbf{f}}_t = \Delta( \boldsymbol{\xi}_t, \mathbf{\Phi}_t ) \quad ,
\label{eqn::csforbs_compressivebackgroundsubtraction}
\end{align}
where $\boldsymbol{\xi}_t = \mathbf{y}_t - \boldsymbol{\beta}_t$ and $\boldsymbol{\beta}_t = \mathbf{\Phi}_t\mathbf{b}$.

As we will discuss in Section \ref{secn::sensingmatrixdesign}, we construct $\mathbf{\Phi}_t$ by taking a subset of rows from a fixed $N^2 \times N^2$ matrix, $\mathbf{\Phi}$, and rescaling the result.  We can therefore calculate $\boldsymbol{\beta}_t$ from $\boldsymbol{\beta} = \mathbf{\Phi}\mathbf{b}$ by similarly dropping components and rescaling.  Noting (\ref{eqn::backgroundsubtraction_foregroundmodel}), a maximum-likelihood estimate of $\boldsymbol{\beta}$ can be found by computing the mean of compressive measurements of a background-only video sequence, i.e.,
\begin{align}
\boldsymbol{\beta} = \frac{1}{J} \sum_{j=1}^J \mathbf{y}_j \quad ,
\end{align}
where $\mathbf{y}_j = \mathbf{\Phi}\mathbf{x}_j$ and $\vert \mathcal{F}_j \vert = 0$ for all $j$ in the summation.  These measurements can be obtained in advance by using the full sensing matrix, $\mathbf{\Phi}$, to observe the scene when it is known that there is no foreground component.
\section{Sensing Matrix Design}
\label{secn::sensingmatrixdesign}
In this section, we will discuss our method for constructing adaptive rate measurement matrices for the purpose of recovering sparse signals from a minimal amount of measurements.

\subsection{Theoretical Guarantees}
In Section \ref{secn::cs}, we presented a theoretical result from CS literature that states that $\Delta$ will exactly recover an $s$-sparse $\mathbf{f}$ from $\boldsymbol{\xi}$ if $\mathbf{\Phi}$ exhibits the RIP of order $2s$ with $\delta_{2s} \leq \sqrt{2}-1$.  One of the most prevalent methods discussed in the literature for constructing such matrices involves drawing each matrix entry from a Gaussian distribution with parameters that depend on the number of rows that the matrix possesses.  For $\mathbf{\Phi} \in \mathbb{R}^{M \times N^2}$, this technique defines entries $\phi_{ij}$ as independent realizations of a Gaussian random variable with zero mean and variance $1/M$, i.e.,
\begin{align}
\phi_{ij} \sim \mathcal{N}(0,1/M) \quad .
\label{eqn::sensingmatrixdesign_gaussianmeasurementmatrix}
\end{align}
Baraniuk \emph{et al.} \cite{Baraniuk2008} provide the following theoretical result for this construction technique: for a given $\delta \in (0,1)$ and positive integers $M$ and $s$, $\mathbf{\Phi} \in \mathbb{R}^{M \times N^2}$ constructed according to (\ref{eqn::sensingmatrixdesign_gaussianmeasurementmatrix}) exhibits the RIP of order $s$ with $\delta_s = \delta$ with probability exceeding
\begin{align}
1 - 2e^{ -c_0(\delta /2)M + s\left( \log(eN^2/s)+\log(12/\delta) \right) } \quad ,
\label{eqn::sensingmatrixdesign_gaussianconstructionguarantee}
\end{align}
where $c_0(x) = x^2/4 - x^3/6$.

The scenarios discussed in this paper require us to find the minimum $M$ that will ensure the constructed matrix can successfully recover $s$-sparse signals.  Therefore, we now consider the case where $\delta$, $s$, and $N^2$ are fixed.  If we impose a lower bound, $\tau_g$, on the probability of success given by (\ref{eqn::sensingmatrixdesign_gaussianconstructionguarantee}), rearranging terms reveals that the theory requires
\begin{align}
M \geq \frac{s[1+\log(\frac{N^2}{s})+\log(\frac{12}{\delta})]+\log(\frac{2}{1-\tau_g})}{\frac{\delta^{2}}{16}(1-\frac{\delta}{3})} \quad .
\label{eqn::sensingmatrixdesign_mlowerbound}
\end{align}

For practical measurement matrices, we are only interested in the case where $N^2 \geq M$ (i.e., matrices for which compression actually occurs).  Combining this requirement with (\ref{eqn::sensingmatrixdesign_mlowerbound}) yields the following lower bound for $N^2/s$:
\begin{align}
\frac{N^2}{s} \geq \frac{ \log(\frac{N^2}{s})+ \frac{1}{s}\log(\frac{2}{1-\tau_g}) }{\frac{\delta^{2}}{16}(1-\frac{\delta}{3})} + \frac{1+\log(\frac{12}{\delta})}{\frac{\delta^{2}}{16}(1-\frac{\delta}{3})} .
\label{eqn::sensingmatrixdesign_sparsityratiolowerbound}
\end{align}
For $s$-sparse signals, the reconstruction guarantee that accompanies $\Delta$ requires that $\mathbf{\Phi}$ exhibits the RIP of order $2s$ with $\delta_{2s} \leq \sqrt{2}-1$.  Using only the second term of the lower bound in (\ref{eqn::sensingmatrixdesign_sparsityratiolowerbound}) and noting that the first term is always positive, we see that requiring such a $\delta_{2s}$ means that $s/N^2$ can be no greater than $\sim 0.0011$.

In our system, $s/N^2$ represents the percentage of foreground pixels in the image, and it is unreasonable to expect that this quantity will never exceed $0.11\%$.  Therefore, if we wish to use CS for compression (i.e., with a measurement matrix that has fewer rows than columns), we must design and use matrices without the guarantee provided by the above result.  However, that result is merely sufficient: in the next part, we will experimentally show that similarly-constructed matrices with far fewer rows are indeed still able to provide measurements that enable accurate sparse signal reconstruction.

\subsection{Practical Sensing Matrix Design Based on Phase Diagrams}
Given a candidate sensing matrix construction technique, Donoho and Tanner \cite{Donoho2010} discuss an associated \emph{phase diagram}: a numerical representation of how useful the generated matrices are for CS.  Specifically, the ratios $M/N^2$ (signal undersampling) and $s/M$ (signal sparsity) are considered.  A phase diagram is a function defined over the \emph{phase space} $(M/N^2,s/M) \in \left[ 0,1 \right]^2$.  We discretize this space and perform multiple sense-and-reconstruct experiments at each grid point in order to approximate the phase diagram there: the value of $M/N^2$ provides the information necessary for matrix construction, and $s/M$ provides the information necessary to generate random sparse signals.  We make the approximation using the percentage of trials that result in successful signal recovery, which we define as a normalized $\ell_2$ reconstruction error of $10^{-3}$ or less.

Even though we cannot use the theoretical guarantee discussed earlier in this section, the first matrix construction technique we use is based on randomly-generated matrices that rely on independent realizations of a Gaussian random variable. Specifically, we use the following construction technique: we generate $\mathbf{\Phi} \in \mathbb{R}^{N^2 \times N^2}$ by drawing each entry according to (\ref{eqn::sensingmatrixdesign_gaussianmeasurementmatrix}).  Then, for a given value of $M_t$, we form the corresponding $M_t \times N^2$ matrix $\mathbf{\Phi}_t$ via
\begin{align}
\mathbf{\Phi}_t = \sqrt{\frac{N^2}{M_t}} \; \mathbf{\Phi}_{1:M_t} \quad ,
\label{eqn::sensingmatrixdesign_phitformation}
\end{align}
where $\mathbf{\Phi}_{1:M_t}$ denotes the submatrix of $\mathbf{\Phi}$ corresponding to the first $M_t$ rows.  The scaling factor ensures that the relationship between the variance and the number of rows defined in (\ref{eqn::sensingmatrixdesign_gaussianmeasurementmatrix}) is preserved.

We also analyze a second matrix construction technique based on the discrete Fourier transform (DFT).  Specifically, we generate $\mathbf{\Phi} \in \mathbb{C}^{N^2 \times N^2}$ by randomly permuting the rows of the DFT matrix and form $\mathbf{\Phi}_t$ according to (\ref{eqn::sensingmatrixdesign_phitformation}).

\begin{figure*}
	\centering
	\subfloat[Gaussian]{\includegraphics[width=0.48\textwidth]{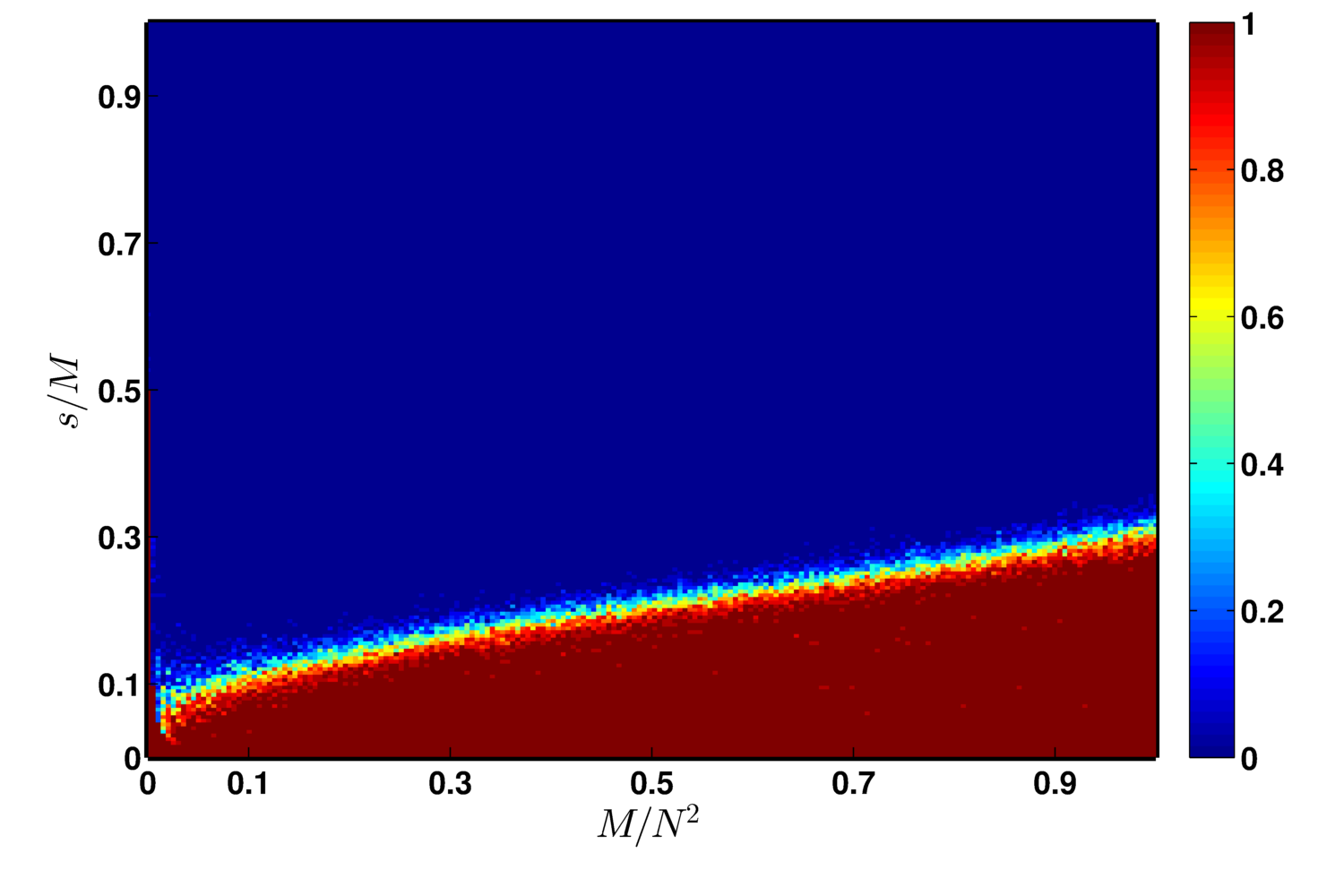} \label{fig::sensingmatrixdesign_phasediagram_gaussian}}
	\subfloat[Fourier]{\includegraphics[width=0.48\textwidth]{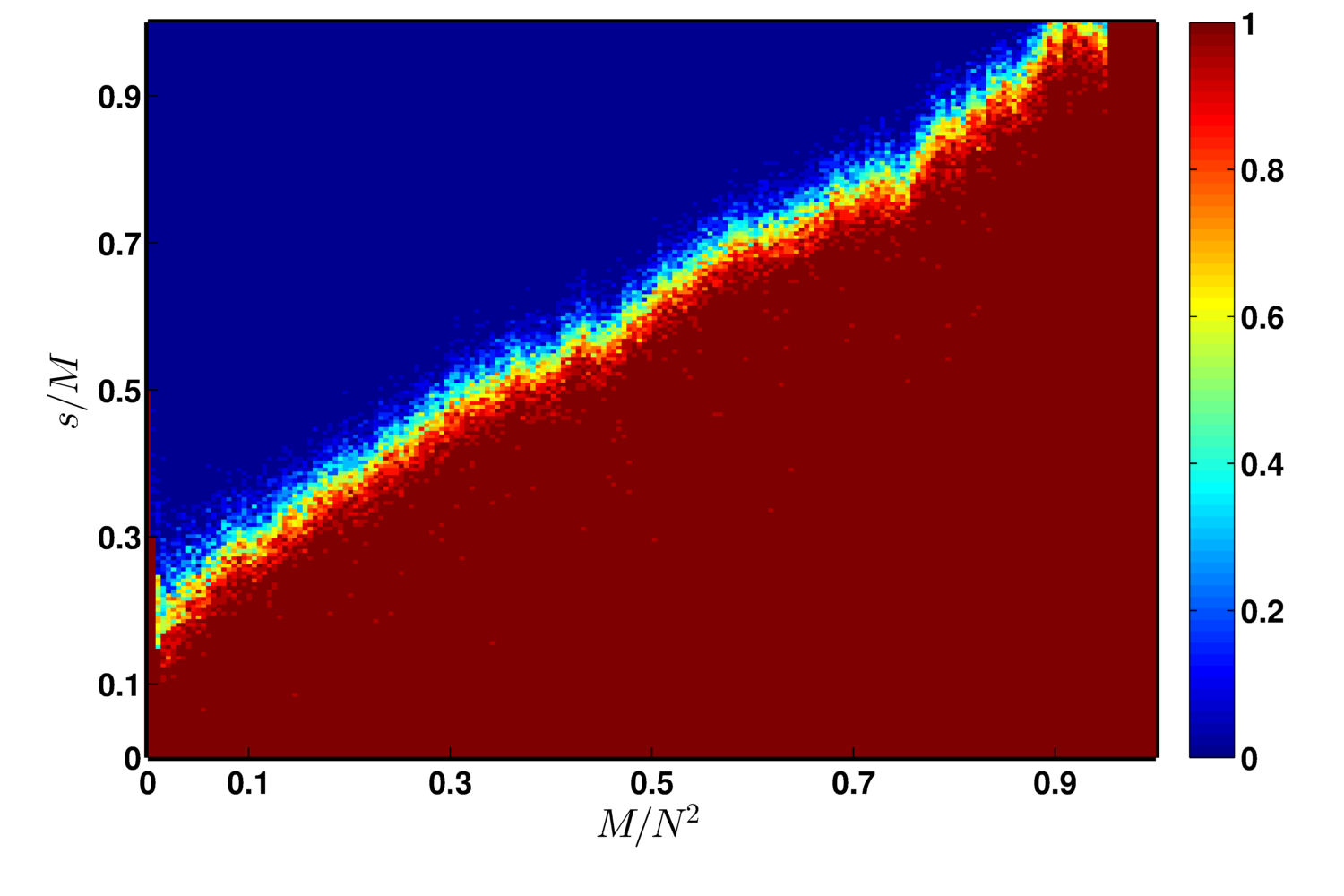} \label{fig::sensingmatrixdesign_phasediagram_fourier}}
	\caption{\footnotesize{Phase diagrams for Gaussian and Fourier measurement ensembles.  Color corresponds to probability of successful reconstruction (here, normalized $\ell_2$ error below $10^{-3}$).}}
	\label{fig::sensingmatrixdesign_phasediagrams}
\end{figure*}

In this paper, we will make predictions regarding the sparsity of the signals we are about to observe.  Given a prediction $s_t$, we will seek the minimum $M_t$ such that  (\ref{eqn::sensingmatrixdesign_phitformation}) generates a sensing matrix capable of providing enough measurements to ensure accurate reconstruction of $s_t$-sparse signals.  In order to determine the mapping from $s_t$ to $M_t$, we use the associated phase diagram.  We construct this diagram (see Figure \ref{fig::sensingmatrixdesign_phasediagrams}) during a one-time, offline analysis.  Then, given $s_t$ and a minimum probability of reconstruction success $\tau_d \in (0,1)$, we use the phase diagram as a lookup table to find the smallest value of $M_t$ that yields at least a $\tau_d$ success rate for $s_t$-sparse signals.

\section{Method I: Cross Validation}
\label{secn::method1}
In this section, we describe a rate-adaptive CS method that utilizes a set of linear cross-validation measurements $\boldsymbol{\chi}_t = \mathbf{\Psi}\mathbf{x}_t$.  An earlier version of this work was presented by Warnell \emph{et al.} \cite{Warnell2012}.

\subsection{Compressive Sensing with Cross Validation}
\label{secn::method1_cswithcv}
Let $\boldsymbol{\xi}_t \in \mathbb{C}^{M_t}$ be a set of compressive measurements of a sparse signal $\mathbf{f}_t \in \mathbb{R}^{N^2}$ obtained using $\mathbf{\Phi}_t$, i.e., $\boldsymbol{\xi}_t = \mathbf{\Phi}_t\mathbf{f}_t$.  In this section, we will use $\hat{\mathbf{f}}_t^{(s)}$ to denote the $s$-sparse point estimate of this signal obtained using $\Delta(\boldsymbol{\xi}_t,\mathbf{\Phi}_t)^{(s)}$, where $\Delta$ is defined as in (\ref{eqn::cs_l1decoding}) and $\cdot^{(s)}$ denotes a truncation operation that sets all but the $s$ largest-magnitude components of the vector-valued argument to zero.

Ward \cite{Ward2009} bounds the error of the above estimate using a cross-validation technique that is based on the Johnson-Lindenstrauss lemma \cite{Johnson1984}.  At the same time $\boldsymbol{\xi}_t$ is collected, we use a static \emph{cross-validation matrix} $\mathbf{\Psi} \in \mathbb{C}^{r \times N^2}$ to collect \emph{cross-validation measurements} $\boldsymbol{\gamma}_t = \mathbf{\Psi} \mathbf{f}_t$.  We construct $\boldsymbol{\Psi}$ by drawing each of its entries from an i.i.d. Bernoulli distribution with zero mean and variance $1/r$.  Such a construction leads to the following statement: for given accuracy and confidence parameters $\epsilon$ and $\rho$ (respectively), $r \geq 8\epsilon^{-2}\log\frac{1}{2\rho}$ rows suffice to ensure that
\begin{align}
(1-\epsilon)^2 \leq \frac{\lVert \mathbf{f}_t - \hat{\mathbf{f}}_t^{(s)} \rVert_2^2}{\lVert \boldsymbol{\gamma}_t - \boldsymbol{\Psi}\hat{\mathbf{f}}_t^{(s)} \rVert_2^2} \leq (1+\epsilon)^2
\label{eqn::method1_cvbound}
\end{align}
with probability exceeding $1-\rho$.

Let $e_s(\mathbf{f}_t)_p$ denote the optimal $s$-sparse approximation error measured with respect to the $\ell_p$ norm, i.e.,
\begin{align}
e_s(\mathbf{f}_t)_p = \argmin_{\lVert \mathbf{z} \rVert_0 \leq s} \; \lVert \mathbf{f}_t - \mathbf{z} \rVert_p \quad ,
\end{align}
where the $\ell_p$-norm is given by $\lVert \mathbf{x} \rVert_p = \left( \sum_i \vert x(i) \vert^p \right)^{1/p}$.
Using the fact that $\hat{\mathbf{f}}_t^{(s)}$ is $s$-sparse, the upper bound in (\ref{eqn::method1_cvbound}) can be extended to $e_s(\mathbf{f}_t)_2^2$ as follows:
\begin{align}
e_{\hat{s}_t}(\mathbf{f}_t)_2^2 \leq \lVert \mathbf{f}_t - \hat{\mathbf{f}}_t^{(s)} \rVert_2^2 \leq (1+\epsilon)^2 \lVert \boldsymbol{\gamma_t} - \mathbf{\Psi}\hat{\mathbf{f}}_t \rVert_2^2 \quad .
\label{eqn::method1_cvupperboundoptimal}
\end{align}
That is, the observable CV error can be used to upper bound the unobservable optimal $s$-sparse approximation error.

\subsection{Adaptive-Rate Compressive Sensing via Cross Validation}
Let $s_t$ denote the true value of the foreground sparsity at time $t$, i.e., $s_t = \vert \mathcal{F}_t \vert$.  The method we present here relies on an estimate of this quantity, which we denote as $\hat{s}_t$.  Before sensing begins at time $t$, we assume $\mathbf{f}_t$ to be $\hat{s}_t$-sparse, and select the corresponding minimal $M_t$ (and thus $\mathbf{\Phi}_t$) according to the phase diagram technique described in Section \ref{secn::sensingmatrixdesign}.  We then use $\mathbf{\Phi}_t$ and $\mathbf{\Psi}$ to collect $\mathbf{y}_t$ and $\boldsymbol{\chi}_t$.  Using the technique described in Section \ref{secn::backgroundsubtraction}, we can find $\boldsymbol{\xi}_t$ and form the foreground estimate $\hat{\mathbf{f}}_t^{(\hat{s}_t)}$.  In a similar fashion, we can also find $\boldsymbol{\gamma}_t$ by subtracting a precalculated set of cross-validation measurements of the static signal component, $\boldsymbol{\zeta} = \mathbf{\Psi}\mathbf{b}$, from $\boldsymbol{\chi}_t$.  Finally, we select $\hat{s}_{t+1}$ based on the result of a multiple hypothesis test that uses $\boldsymbol{\gamma}_t$ and $\hat{\mathbf{f}}_t^{(\hat{s}_t)}$.

We formulate the multiple hypothesis test by first assuming that we are able to observe $e_{\hat{s}_t}(\mathbf{f}_t)_2^2$.  We define the null hypothesis, $\mathbf{H}_0$, as the scenario under which $\hat{s}_t$ exceeds $s_t$.  If this hypothesis is true, then $\mathbf{f}_t^{(\hat{s}_t)}$ (i.e., the optimal $\hat{s}_t$-sparse approximation to $\mathbf{f}_t$) captures all $s_t$ foreground pixels and $(\hat{s}_t - s_t)$ background pixels while neglecting the remaining $(N-\hat{s}_t)$ background pixels.  Using (\ref{eqn::backgroundsubtraction_foregroundmodel}), it can be shown that $e_{\hat{s}_t}(\mathbf{f}_t)_2^2$ is a random variable with mean, $\mu_0$, and variance, $\sigma_0^2$, given by
\begin{align}
\mu_0 &= (N-\hat{s}_t)\sigma_b^2 \nonumber \\
\sigma_0^2 &= 2(N-\hat{s}_t)\sigma_b^4 \quad .
\label{eqn::method1_hypothesis0meanvariance}
\end{align}

We also define a set of hypotheses that are possible when $\mathbf{H}_0$ is not true.  Let $\mathbf{H}_k, k \in \{ \hat{s}_t+1, \ldots, N \}$ describe the scenario under which $s_t = k$.  Under $\mathbf{H}_k$, $\mathbf{f}_t^{(\hat{s}_t)}$ cannot capture all $k$ foreground pixels: it neglects the smallest $(k-\hat{s}_t)$ of them and the $(N-k)$ background pixels.  Using (\ref{eqn::backgroundsubtraction_foregroundmodel}), it can be shown that the mean, $\mu_k$, and variance, $\sigma_k^2$, of $e_{\hat{s}_t}(\mathbf{f}_t)_2^2$ under these hypotheses are given by

{\small
\begin{align}
\mu_k = & (N-k)\sigma_b^2 +  \frac{1}{3}(k-\hat{s}_t)(\tau^2 + \tau + 1) & \nonumber \\
\sigma_k^2 = & \frac{1}{9} \left[ (k-\hat{s}_t)^2 - (k-\hat{s}_t) \right] (\tau^2+\tau+1)^2 & \nonumber \\
+ &\frac{1}{5} (k-\hat{s}_t)(\tau^4+\tau^3+\tau^2+\tau+1) & \nonumber \\
+ &\left[ (N-k)^2 + 2(N-k) \right] \sigma_b^4 &  \nonumber \\
+ &\frac{2}{3}(N-k)(k-\hat{s}_t)(\tau^2+\tau+1)\sigma_b^2 - \mu_k^2 \quad . &
\label{eqn::method1_hypothesiskmeanvariance}
\end{align}
}

The hypothesis test can be succintly written as
\begin{align}
\mathbf{H}_0 & : s_t < \hat{s}_t \nonumber \\
\mathbf{H}_k & : s_t = k
\label{eqn::method1_idealhypothesistest}
\end{align}
for $k \in \left\{ \hat{s}_t+1,\ldots,N \right\}$.  Let $q_k$ denote the probability density function for $e_{\hat{s}_t}(\mathbf{f}_t)_2^2$ under the assumption that $\mathbf{H}_k$ is true for $k \in \left\{ 0,\hat{s}_t+1,\ldots,N \right\}$.  We will evaluate explicit assumptions regarding the form of $q_k$ in Section \ref{secn::experiments}.  The optimal decision rule for (\ref{eqn::method1_idealhypothesistest}) under the minimum probability of error criterion with an equal prior for each hypothesis is given by
\begin{align}
k^* = \argmax_{k \in \{0,\hat{s}+1,\ldots,N\}} q_k \left( e_{\hat{s}_t}(\mathbf{f}_t)_2^2 \right) \quad .
\label{eqn::method1_idealdecisionrule}
\end{align}

Assuming that the sparsity of $\mathbf{f}_t$ is a slowly-varying quantity, we choose to set $\hat{s}_{t+1}$ equal to what we believe $s_t$ to be.  If $k^*=0$, it is our belief that $\hat{s}_t > s_t$, and we expect that the error in $\hat{\mathbf{f}}_t^{(\hat{s}_t)}$ to be very small.  Therefore, we find the set of foreground entries for this signal,  $\hat{\mathcal{F}}_t = \{ i \; : \; \vert \hat{f}_t^{(\hat{s}_t)}(i) \vert \geq \tau \}$, and set $\hat{s}_{t+1} = \vert \hat{\mathcal{F}}_t \vert$.  For any other value of $k^*$, we set $\hat{s}_{t+1}=k^*$.

Unfortunately, it is impossible to directly observe $e_{\hat{s}_t}(\mathbf{f}_t)_2^2$.  However, we can upper bound this quantity using the cross-validation measurements as specified in (\ref{eqn::method1_cvupperboundoptimal}).  Therefore, we propose the following modification to (\ref{eqn::method1_idealdecisionrule}):
\begin{align}
k^* = \argmax_{k \in \{0,\hat{s}+1,\ldots,N\}} q_k \left( (1+\epsilon)^2 \lVert \boldsymbol{\gamma}_t - \mathbf{\Psi}\hat{\mathbf{f}}_t^{(\hat{s}_t)} \rVert_2^2 \right) \quad .
\label{eqn::method1_cvdecisionrule}
\end{align}

Observing that $\mu_k$ and $\sigma_k^2$ are increasing functions of $k$, it is apparent that (\ref{eqn::method1_cvdecisionrule}) will potentially yield a value of $k^*$ greater than that which would have been selected by (\ref{eqn::method1_idealdecisionrule}).  This will result in a higher-than-necessary measurement rate at time $t+1$, but it will not negatively impact the quality of $\hat{\mathbf{f}}_{t+1}^{(\hat{s}_{t+1})}$.

We term the strategy we have outlined above \emph{adaptive-rate compressive sensing via cross validation (ARCS-CV)} and summarize the procedure in Algorithm \ref{alg::method1_arcscvalg}.

\begin{algorithm}
\caption{ARCS-CV for Background Subtraction}
\begin{algorithmic}
\label{alg::method1_arcscvalg}
\REQUIRE $\mathbf{\Phi}, \mathbf{\Psi}, \hat{s}_t, \boldsymbol{\beta}, \boldsymbol{\zeta}, \sigma_b^2, \tau$ \\
\STATE Select $M_t$ using $\hat{s}_t$ and the phase diagram lookup table  \\
\STATE Form $\mathbf{\Phi_t}$ and $\boldsymbol{\beta}_t$ \\
\STATE Obtain image measurements $\mathbf{y_t}$, $\boldsymbol{\chi}_t$ \\
\STATE Compute foreground-only measurements $\boldsymbol{\xi}_t$, $\boldsymbol{\gamma}_t$ \\
\STATE Estimate foreground: $\mathbf{\hat{f}}_t^{(\hat{s}_t)} = \Delta(\boldsymbol{\xi}_t,\mathbf{\Phi}_t)^{(\hat{s}_t)}$ \\
\STATE Compute $k^*$ using (\ref{eqn::method1_cvdecisionrule})\\
\IF{$k^*=0$}
\STATE $\hat{s}_{t+1} = \vert \hat{\mathcal{F}}_t \vert$
\ELSE
\STATE $\hat{s}_{t+1} = k^*$
\ENDIF
\end{algorithmic}
\end{algorithm}
\section{Method II: Low-Resolution Tracking}
\label{secn::method2}
In this section, we propose an adaptive method that utilizes a much richer form of side information than the random projections of the previous section: low-resolution images, $Z_t$, that have been captured using a traditional (i.e., non-compressive) camera.

\subsection{Low-Resolution Measurements}
We assume that the low- and high-resolution images, $Z_t \in \mathbb{R}^{L \times L}$ and $X_t \in \mathbb{R}^{N \times N} (L<N)$, repectively, are related by a simple downsampling operation.  Let $\mathbf{t}_Z = \begin{bmatrix} t_Z^x & t_Z^y \end{bmatrix}^T$ denote the coordinates of a pixel in the image plane of the low-resolution camera.  If we use $\mathbf{t}_X = \begin{bmatrix} t_X^x & t_X^y \end{bmatrix}^T$ to denote the corresponding coordinate in the image plane of the compressive camera, the effect of the downsampling operation on coordinates is given by
\begin{align}
\mathbf{t}_X = \begin{bmatrix} D & 0 & -\frac{D-1}{2} \\ 0 & D & -\frac{D-1}{2} \end{bmatrix} \begin{bmatrix} \mathbf{t}_Z \\ 1 \end{bmatrix} \quad ,
\label{eqn::method2_downsamplingcoordinatemapping}
\end{align}
where we assume the dowsampling factor, $D=N/L$, to be an integer.  Using (\ref{eqn::method2_downsamplingcoordinatemapping}), each pixel in $Z_t$ maps to the center of a unique $D \times D$ block of pixels in $X_t$.  The effect of the downsampling operation on image intensity is given by averaging the intensities within this block, i.e.,
\begin{align*}
Z_t(\mathbf{t}_Z) = \frac{1}{D^2} \sum_{\mathbf{t}_X \in \mathcal{B}(\mathbf{t}_Z)} X_t(\mathbf{t}_X) \quad ,
\end{align*}
where the coordinates of the pixels in the block are given explicitly as
\begin{align*}
\mathcal{B}(\mathbf{t}_Z) = &\left\{ (\mathbf{t}_Z^x-1)D+1,\ldots,\mathbf{t}_Z^xD \right\} \times \nonumber \\
 & \left\{ (\mathbf{t}_Z^y-1)D+1,\ldots,\mathbf{t}_Z^yD \right\} .
\end{align*}

\subsection{Object Tracking and Foreground Sparsity}
\label{secn::method2_objecttracking}
Given $Z_t$, we assume that we are able to track the foreground objects.  Specifically, we assume that at each time index, we are able to estimate a zero-skew affine warp parameter $\mathbf{p}_t = \begin{bmatrix} p_t(1) & \cdots & p_t(4) \end{bmatrix}^T$ that maps coordinates in an object template image, $T$, to their corresponding location in $Z_t$.  Using $\mathbf{t}_T$ to denote a pixel coordinate in $T$, $\mathbf{p}_t$ specifies the corresponding coordinate in $Z_t$ via
\begin{align}
\mathbf{t}_Z = \begin{bmatrix} p_t(1) & 0 & p_t(3) \\ 0 & p_t(2) & p_t(4) \end{bmatrix} \begin{bmatrix} \mathbf{t}_T \\ 1 \end{bmatrix} \quad .
\label{eqn::method2_templatecoordinatemapping}
\end{align}
We further assume that the time-evolution of $\mathbf{p}_t$ is governed by a known Markov dynamical system, i.e.,
\begin{align}
\mathbf{p}_{t} = \mathbf{u}_t \left( \mathbf{p}_{t-1}, \boldsymbol{\eta}_t \right) \quad ,
\label{eqn::method2_trackevolution}
\end{align}
for known $\mathbf{u}_t$ and i.i.d. system noise $\boldsymbol{\eta}_t$.

Let $\left\{ \mathbf{t}_i \; : \; i \in \mathbb{Z}/4\mathbb{Z} \right\}$ be the set of corner coordinates of $T$ in any order that traces its outline.  Then, given $\mathbf{p}_t$, we can calculate the position of the tracked object's bounding box in $F_t$ using (\ref{eqn::method2_templatecoordinatemapping}) and (\ref{eqn::method2_downsamplingcoordinatemapping}).  We shall assume that the area of this bounding box specifies the number of foreground components in $\mathbf{f}_t$, i.e., $s_t$.  If this area is not integer-valued, we simply round up.  Using the well-known formula for the area of a polygon from its corner coordinates, $s_t$ can be written as $s_t = h(\mathbf{p}_t)$, where

\noindent
{\small
\begin{align}
h(\mathbf{p}_t) = \left\lceil \left\vert \frac{D^2[p_t(1)p_t(4)-p_t(2)p_t(3)]}{2} 	\sum_{i \in \mathbb{Z}/4\mathbb{Z}} \mathcal{T}(i) \right\vert \right\rceil ,
\label{eqn::method2_tracktosparsity}
\end{align}
}
and $\mathcal{T}(i) = t_i^x t_{i+1}^y - t_i^y t_{i+1}^x$.  Above, $\lceil \cdot \rceil$ represents the ceiling function.

From (\ref{eqn::method2_tracktosparsity}), it is clear that the distribution of the random variable $s_t$ is a function of the distribution of $\mathbf{p}_t$.  For the remainder of this section, we will use $q_t(s_t) = p(s_t \vert \mathbf{p}_t)$ to denote the corresponding probability mass function.

\begin{figure}
	\centering
	\subfloat[]{\includegraphics[scale=1.0]{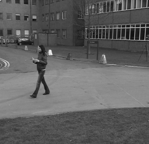} \label{fig::method2_downsampleandtrack_hr}}
	\centering
	\subfloat[]{\includegraphics[scale=1.0]{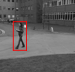} \label{fig::method2_downsampleandtrack_lr}}
	\caption{\footnotesize{Illustration of the downsampling and low-resolution tracking process utilized by ARCS-LRT for a sample image from the \texttt{PETS\_2009} dataset.  (\protect\subref{fig::method2_downsampleandtrack_hr}) corresponds to the high-resolution image for which we seek to perform compressive foreground reconstruction.  (\protect\subref{fig::method2_downsampleandtrack_lr}) corresponds to the low-resolution obtained by the secondary, non-compressive camera.  The bounding box around the woman corresponds to the output of a tracking algorithm.}}
	\label{fig::method2_downsampleandtrack}
\end{figure}

Figure \ref{fig::method2_downsampleandtrack} illustrates the relationship between a typical high- and low-resolution image pair and shows an example bounding box found by a tracker using the low-resolution image.

\subsection{Sparsity Estimation}
\label{secn::method2_sparsityestimation}
We now turn our attention to selecting a value to use for $s_t$, $\hat{s}_t$, on the basis of the previous image's track, $\mathbf{p}_{t-1}$.  Once a value has been selected, we use the method presented in Section \ref{secn::sensingmatrixdesign} to select a minimal $M_t$ and the corresponding $\mathbf{\Phi}_t$.  We then use $\mathbf{\Phi}_t$ to collect compressive measurements of $X_t$ and calculate $\boldsymbol{\xi}_t$.  Using this procedure, the $\Delta$-generated estimate $\hat{\mathbf{f}}_t$ will obey
\begin{align}
\lVert \mathbf{f}_t - \hat{\mathbf{f}}_t \rVert_2 \leq \frac{C_0 e_{\hat{s}_t}(\mathbf{f}_t)_1}{\sqrt{\hat{s}_t}} \quad ,
\label{eqn::method2_l1recoverybound}
\end{align}  
where $e_{\hat{s}_t}(\cdot)_1$ represents the optimal $\hat{s}_t$-sparse $\ell_1$ estimation error \cite{Baraniuk2011a}.  The value of the constant in (\ref{eqn::method2_l1recoverybound}) is given explicitly by
\begin{align*}
&C_0 = \frac{2-(2-\sqrt{2})\delta_{2\hat{s}_t}}{1-(1-\sqrt{2})\delta_{2\hat{s}_t}} \quad .
\end{align*}

One criterion we will consider when selecting $\hat{s}_t$ is the expected value of the $\ell_2$ reconstruction error, i.e., we would like $\hat{s}_t$ to minimize $\mathbb{E} \left\{ \lVert \mathbf{f}_t - \hat{\mathbf{f}}_t \rVert_2 \right\}$.  However, since the nonlinearity of $\Delta$ makes determining the statistics of that quantity very difficult, we instead look to minimize the right-hand side of (\ref{eqn::method2_l1recoverybound}).  It is easy to see that this quantity can be minimized by selecting $\hat{s}_t$ as high as possible, which would provide no compression.  Therefore, inspired by results from the model-order selection literature \cite{Kashyap1977} \cite{Schwarz1978} \cite{Rissanen1978}, we penalize larger values of $\hat{s}_t$ and instead propose to select $\hat{s}_t$ by solving
\begin{align}
\hat{s}_t = \argmin_{\hat{s}} \; \mathbb{E}\left\{ \frac{C_0 e_{\hat{s}}(\mathbf{f}_t)_1}{\sqrt{\hat{s}}} \right\} + \lambda \hat{s} \quad ,
\label{eqn::method2_costfunction}
\end{align}
where $\lambda$ is an importance factor that specifies the tradeoff between low reconstruction error and a small sparsity estimate.

Using the law of total expectation, the foreground model (\ref{eqn::backgroundsubtraction_foregroundmodel}), and techniques similar to those used in Section \ref{secn::method1}, we can rewrite (\ref{eqn::method2_costfunction}) as

\begin{align}
\hat{s}_t = \argmin_{\hat{s}} \frac{C_0}{\sqrt{\hat{s}}} \left[ \mathcal{J}_0(\hat{s}) + \mathcal{J}_1(\hat{s}) \right] + \lambda \hat{s} \quad ,
\label{eqn::method2_expandedcostfunction}
\end{align}
where
\begin{flalign*}
\mathcal{J}_0 = &\sum_{k=1}^{\hat{s}} \sqrt{2/\pi}(N-\hat{s})\sigma_b q_t(k) & \\
\mathcal{J}_1 = &\sum_{k=\hat{s}+1}^N \left[ (k-\hat{s})(1+\tau)/2 + \sqrt{2/\pi}(N-k)\sigma_b \right] q_t(k) . &
\end{flalign*}

We term the strategy that we have outline above as \emph{adaptive-rate compressive sensing via low-resolution tracking (ARCS-LRT)} and summarize the procedure in Algorithm \ref{alg::method2_arcslrt}.

\begin{algorithm}
\caption{ARCS-LRT for Background Subtraction}
\begin{algorithmic}
\label{alg::method2_arcslrt}
\REQUIRE $\mathbf{\Phi}, \hat{s}_t, \boldsymbol{\beta}, \sigma_b^2, \tau, \lambda$ \\
\STATE Select $M_t$ using $\hat{s}_t$ and the phase diagram lookup table  \\
\STATE Form $\mathbf{\Phi_t}$ and $\boldsymbol{\beta}_t$ \\
\STATE Obtain image measurements $\mathbf{y_t}$, $\mathbf{z}_t$ \\
\STATE Compute foreground-only measurements $\boldsymbol{\xi}_t$ \\
\STATE Estimate foreground: $\mathbf{\hat{f}}_t = \Delta(\boldsymbol{\xi}_t,\mathbf{\Phi}_t)$ \\
\STATE Compute low-resolution object track $\mathbf{p}_t$
\STATE Compute $q_{t+1}$ via (\ref{eqn::method2_trackevolution}) and (\ref{eqn::method2_tracktosparsity})
\STATE Compute $\hat{s}_{t+1}$ by solving (\ref{eqn::method2_expandedcostfunction})
\end{algorithmic}
\end{algorithm}

\section{Experiments}
\label{secn::experiments}
\begin{figure}
	\centering
	\subfloat[]{\includegraphics[width=0.14\textwidth]{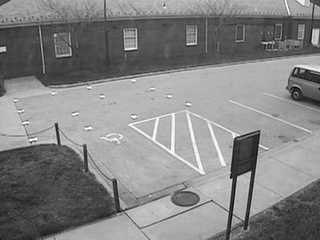} \label{fig::experiments_dataexamples_markercam_background}}
	\subfloat[]{\includegraphics[width=0.14\textwidth]{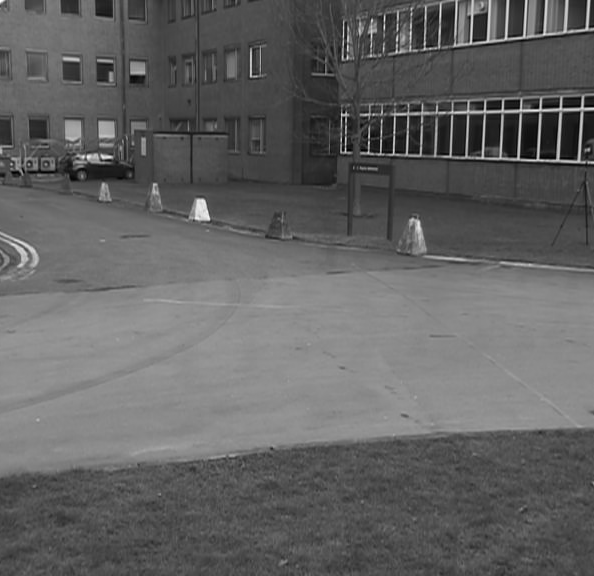} \label{fig::experiments_dataexamples_PETS2009_background}}
	\subfloat[]{\includegraphics[width=0.14\textwidth]{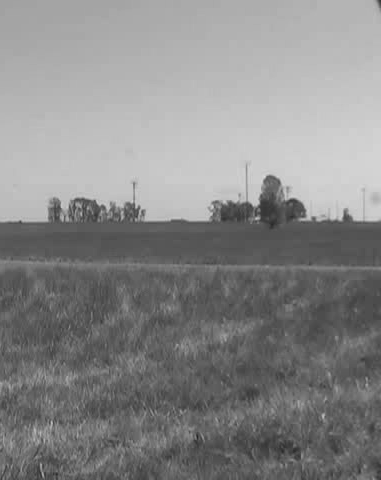} \label{fig::experiments_dataexamples_convoy2_background}}
		
	\vspace{5pt}
	
	\subfloat[]{\includegraphics[width=0.14\textwidth]{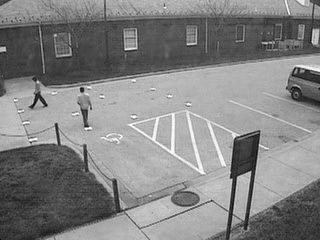} \label{fig::experiments_dataexamples_markercam_img}}
	\subfloat[]{\includegraphics[width=0.14\textwidth]{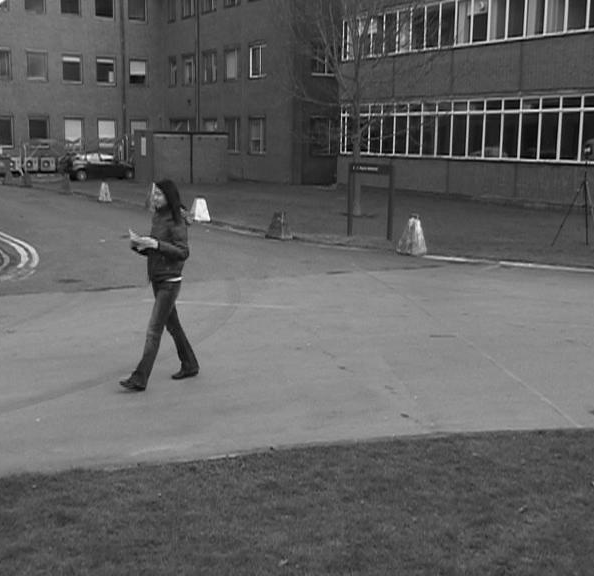} \label{fig::experiments_dataexamples_PETS2009_img}}
	\subfloat[]{\includegraphics[width=0.14\textwidth]{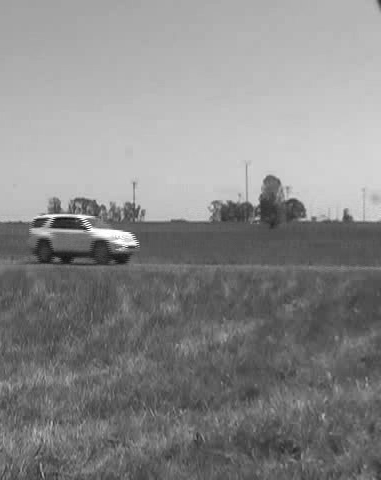} \label{fig::experiments_dataexamples_convoy2_img}}
		
	\vspace{5pt}
	
	\subfloat[]{\includegraphics[width=0.14\textwidth]{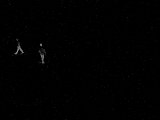} \label{fig::experiments_dataexamples_markercam_foreground}}
	\subfloat[]{\includegraphics[width=0.14\textwidth]{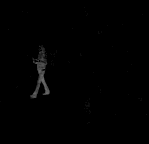} \label{fig::experiments_dataexamples_PETS2009_foreground}}
	\subfloat[]{\includegraphics[width=0.14\textwidth]{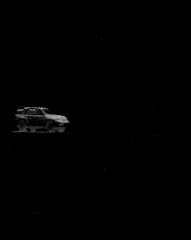} \label{fig::experiments_dataexamples_convoy2_foreground}}
	
	\caption{\footnotesize{Example images from the \texttt{marker\_cam}, \texttt{PETS2009\_S2L1}, and \texttt{convoy2} (columns one, two, and three, respectively), video sequences.  The first row contains the background images, the second row contains an image with both foreground and background components, and the third image contains the corresponding foreground component.}}
	\label{fig::experiments_dataexamples}
\end{figure}

We tested the proposed algorithms on real video sequences captured using traditional cameras.  The compressive, cross-validation, and low-resolution measurements were simulated via software.  The SPGL1 \cite{BergFriedlander:2008} \cite{spgl1:2007} software package was used to implement the decoding procedure (\ref{eqn::cs_l1decoding}).  Three video sequences were used: \texttt{convoy2}, \texttt{marker\_cam}, and \texttt{PETS2009\_S2L1}.  \texttt{convoy2} is a video of vehicles driving past a stationary camera.  The vehicles comprise the foreground, and the foreground sparsity varies as a result of these vehicles sequentially entering and exiting the camera's field of view.  \texttt{marker\_cam} is a video sequence we captured using a surveillance camera mounted to the side of our building at the University of Maryland, College Park.  The sequence begins with a single pedestrian walking in a parking lot, with a second pedestrian joining him halfway through the sequence.  The two pedestrians comprise the foreground, and the foreground sparsity varies due to the entrance of the second pedestrian and the variation in each pedestrian's appearance as he moves relative to the camera.  The \texttt{PETS2009\_S2L1} video sequence is a segment taken from the PETS 2009 benchmark data \cite{PETS2009}.  This sequence consists of four pedestrians entering and exiting the camera's field of view.  Similar to \texttt{marker\_cam}, the foreground sparsity changes as a function of the number and appearance of pedestrians.  Example images from each dataset are shown in Figure \ref{fig::experiments_dataexamples}.

\subsection{Practical Considerations}
Implementation of the ARCS methods presented in Sections \ref{secn::method1} and \ref{secn::method2} requires certain practical choices.  In this part, we describe the choices we made that generated the results presented later in this section.  Specific choices for parameter values for each video sequence are given in Table \ref{table::experiments_parametervalues}.

\begin{table}
\footnotesize
\centering
\caption{Parameter values used in experiments}
\begin{tabular}{|l|c|c|c|c|}
\hline
 & $\sigma_b^2$ & $\tau$ & $\boldsymbol{\Sigma}$ & $\lambda$ \\ 
\hline
\texttt{convoy2} & $\frac{4}{255}^2$ & $0.1$ & {\tiny $\texttt{diag}([1.0 \; 1.0 \; 3.0 \; 3.0])$} & $0.045$ \\
\hline
\texttt{marker\_cam} & $\frac{4}{255}^2$ & $0.1$ & {\tiny $\texttt{diag}([1.0 \; 1.0 \; 3.0 \; 3.0])$} & $1.5$ \\
\hline
\texttt{PETS2009\_S2L1} & $\frac{4}{255}^2$ & $0.1$ & {\tiny $\texttt{diag}([1.0 \; 1.0 \; 3.0 \; 3.0])$} & $0.15$  \\
\hline
\end{tabular}
\label{table::experiments_parametervalues}
\end{table}

\subsubsection{Foreground Model}
The foreground model specified in (\ref{eqn::backgroundsubtraction_foregroundmodel}) is parameterized by $\sigma_b^2$ and $\tau$.  The value that should be used for $\sigma_b^2$ will depend on the quality of the estimate of $\mathbf{b}$ (or, more accurately, $\boldsymbol{\beta}$ in our system): the better (\ref{eqn::backgroundsubtraction_signalmodel}) describes images in the video sequence, the smaller $\sigma_b^2$ can be.  Since $\tau$ represents the foreground-background intensity threshold, its value depends on the value selected for $\sigma_b^2$: $\tau$ should be set high enough to ensure that $\mathcal{N}(\tau;0,\sigma_b^2)$ is sufficiently low, but low enough to ensure that it does not neglect intensities belonging to the foreground.

\subsubsection{ARCS-CV}
The ARCS-CV algorithm developed in Section \ref{secn::method1} relies on the hypothesis test specified in (\ref{eqn::method1_idealhypothesistest}).  While we are able to calculate the first- and second-order moments of $s_t$ under the various hypotheses, the maximum-likelihood decision rule (\ref{eqn::method1_cvdecisionrule}) requires the entire probability density functions, $q_k$, for each.  In our implementation, we approximate these densities by a normal distribution with mean and covariance specified by (\ref{eqn::method1_hypothesis0meanvariance}) and (\ref{eqn::method1_hypothesiskmeanvariance}) under $\mathbf{H}_0$ and $\mathbf{H}_k$, respectively.  That is, we make the approximation $q_k \approx \mathcal{N}(\mu_k,\sigma_k^2)$.  As a consequence of this approximation, we observed that (\ref{eqn::method1_cvdecisionrule}) sometimes yielded a nonzero $k^*$ for sufficiently small cross-validation error upper bounds.  However, when this upper bound is low, it is clear that we should select $\mathbf{H}_0$.  Therefore, we explicitly impose a selection of $\mathbf{H}_0$ for cross-validation error upper bounds that are less than $\mu_0$ by using
\begin{align}
k^{**} = \begin{cases} 0, & (1+\epsilon)^2 \lVert \boldsymbol{\gamma}_t - \mathbf{\Psi}\hat{\mathbf{f}}_t^{(\hat{s}_t)} \rVert_2^2 < \mu_0 \\ k^* , & (1+\epsilon)^2 \lVert \boldsymbol{\gamma}_t - \mathbf{\Psi}\hat{\mathbf{f}}_t^{(\hat{s}_t)} \rVert_2^2 \geq \mu_0 \end{cases}
\end{align}
in place of (\ref{eqn::method1_cvdecisionrule}) in Algorithm \ref{alg::method1_arcscvalg}, where $k^*$ represents the value obtained from (\ref{eqn::method1_cvdecisionrule}).

\subsubsection{ARCS-LRT}
The ARCS-LRT method of Section \ref{secn::method1} requires low-resolution object tracks in order to reason about the sparsity of the high-resolution foreground.  In order to focus on the performance of the adaptive algorithm, we first determined these tracks manually, i.e., by hand-marking bounding boxes around each low-resolution foreground image.  We only did this for images in which the object was fully visible.  We shall also consider automatically-obtained tracks later in this section.

We used $\mathbf{u}_t(\mathbf{p}_{t-1},\boldsymbol{\eta_t}) = \mathbf{p}_{t-1} + \boldsymbol{\eta_t}$ to define the system dynamics in (\ref{eqn::method2_trackevolution}) with $\boldsymbol{\eta_t} \sim \mathcal{N}(\mathbf{0},\boldsymbol{\Sigma})$ i.i.d. for each $t$, where the value of $\Sigma$ should vary with the expected type of object motion.

Given this selection for $\mathbf{u}_t$, $p( \mathbf{p}_t \vert \mathbf{p}_{t-1}) = \mathcal{N}(\mathbf{p}_t; \mathbf{p}_{t-1}, \Sigma)$ represents our belief about the next track given the current one.  Due to the complexity of $h$ in (\ref{eqn::method2_tracktosparsity}), it is difficult to obtain an exact form for $p(s_t \vert \mathbf{p}_{t-1})$.  Therefore, we used the unscented transformation \cite{Julier1985} to obtain the first- and second-order moments, $\mu_{t+1}$ and $\sigma_{t+1}^2$, respectively.  We then approximated $p(s_t \vert \mathbf{p}_{t-1})$ using the pdf for a discrete approximation to the normal distribution with the computed mean and covariance.

The sparsity estimator (\ref{eqn::method2_expandedcostfunction}) requires values for both $C_0$ and $\lambda$.  Since our phase diagram lookup table returns an $M_t$ for which $\Delta$ recovers $\hat{s}_t$-sparse signals, we selected $\delta = 1/4 < \sqrt{2}-1$.  We then selected a $\lambda$ that provided a good balance between the reconstruction error and foreground sparsity.  For each video sequence, we chose this value by trying out many and selecting one that provided a good balance between low reconstruction error and a low sparsity estimate.

Finally, we must compute a solution to (\ref{eqn::method2_expandedcostfunction}).  To do so, we used MATLAB's \texttt{fminbmd} function, which is based on golden selection search and parabolic interpolation \cite{MATLAB:2013}.

\subsection{Comparitive Results}
\begin{figure*}
\centering
	\subfloat[]{\includegraphics[width=0.32\textwidth]{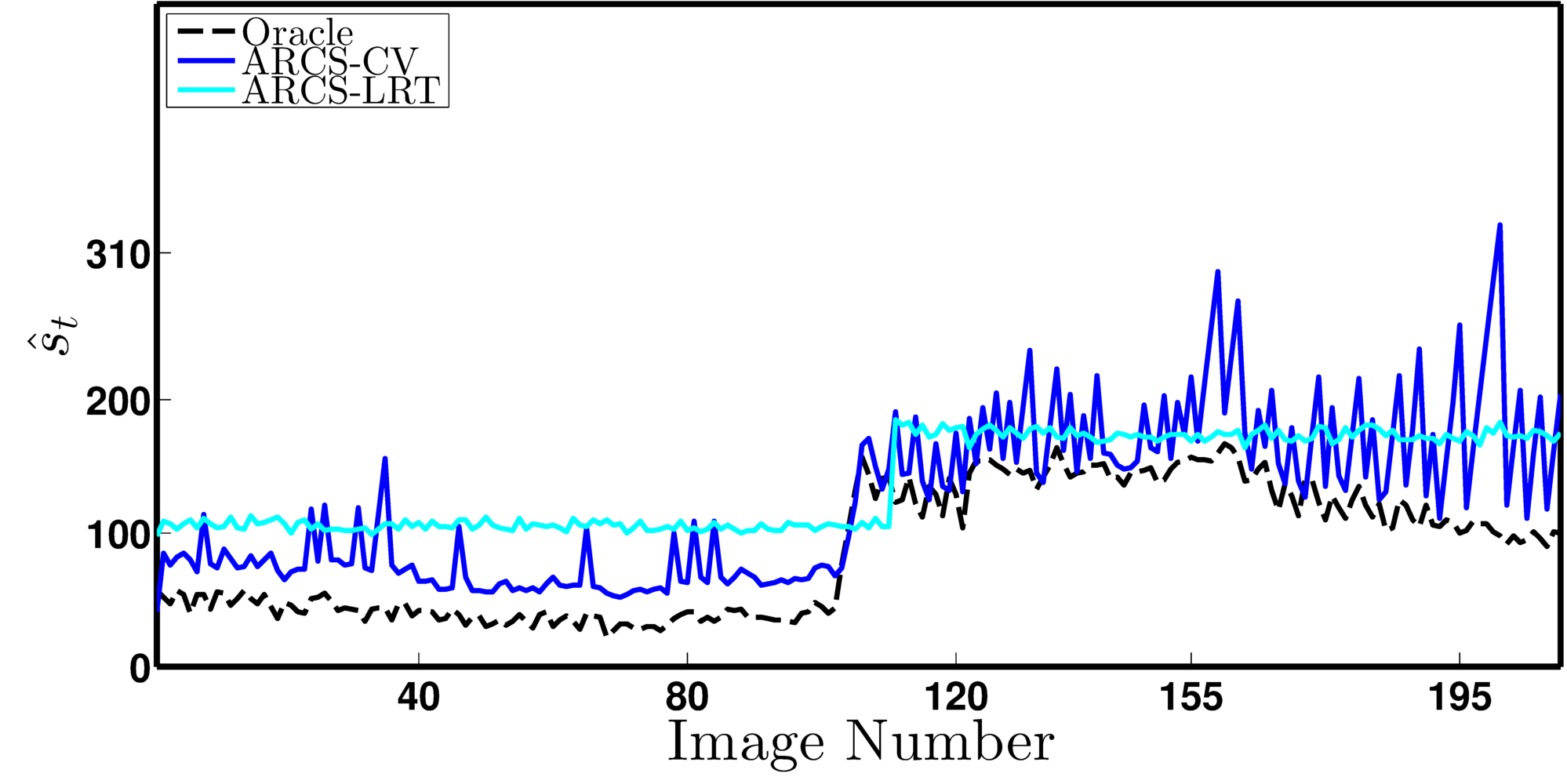} \label{fig::experiments_results_markercam_s}}
	\subfloat[]{\includegraphics[width=0.32\textwidth]{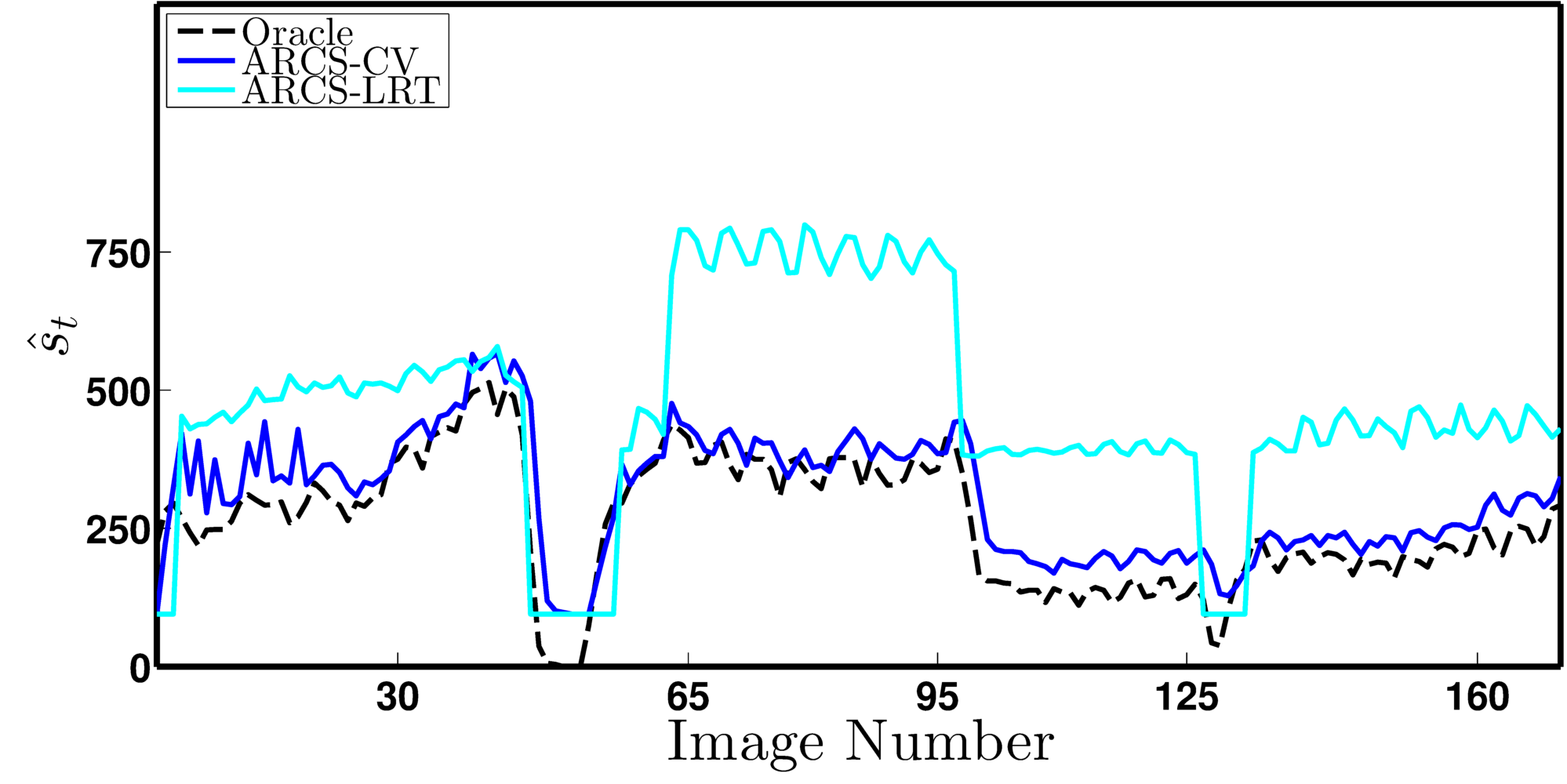} \label{fig::experiments_results_PETS2009_s}}
	\subfloat[]{\includegraphics[width=0.32\textwidth]{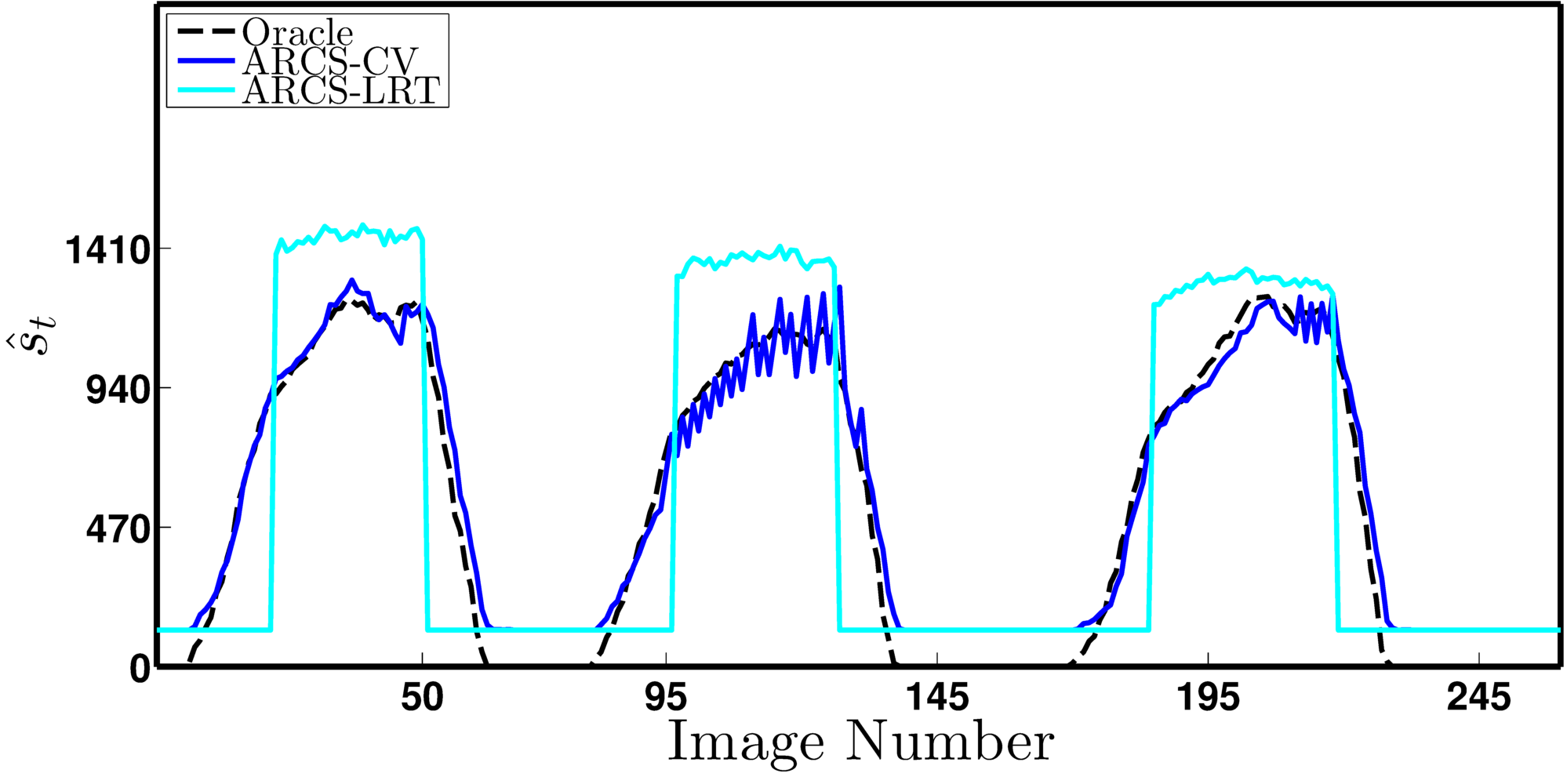} \label{fig::experiments_results_convoy2_s}}
 	
  	\vspace{10pt}
  	
  	\subfloat[]{\includegraphics[width=0.32\textwidth]{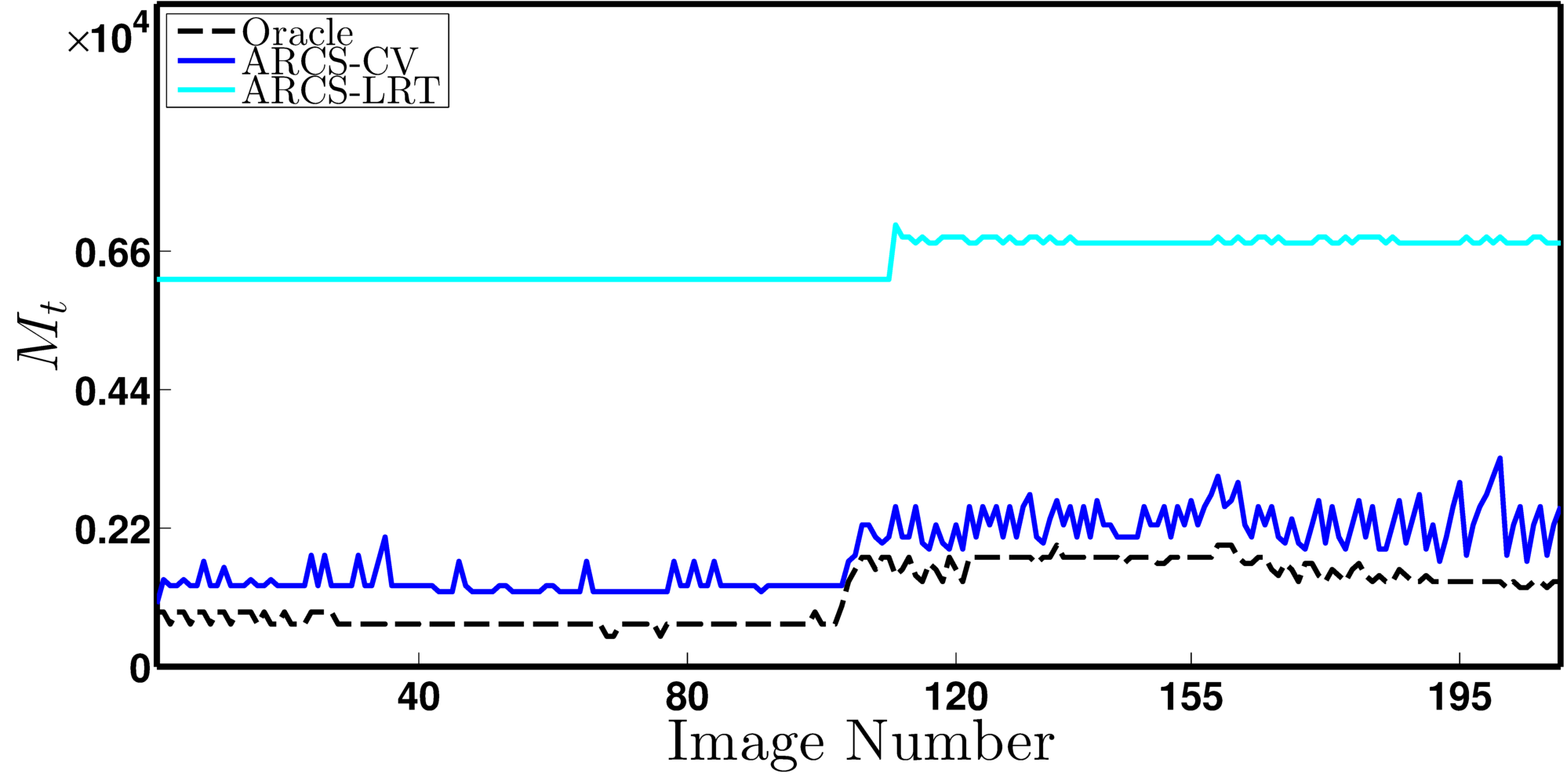} \label{fig::experiments_results_markercam_M}}
  	\subfloat[]{\includegraphics[width=0.32\textwidth]{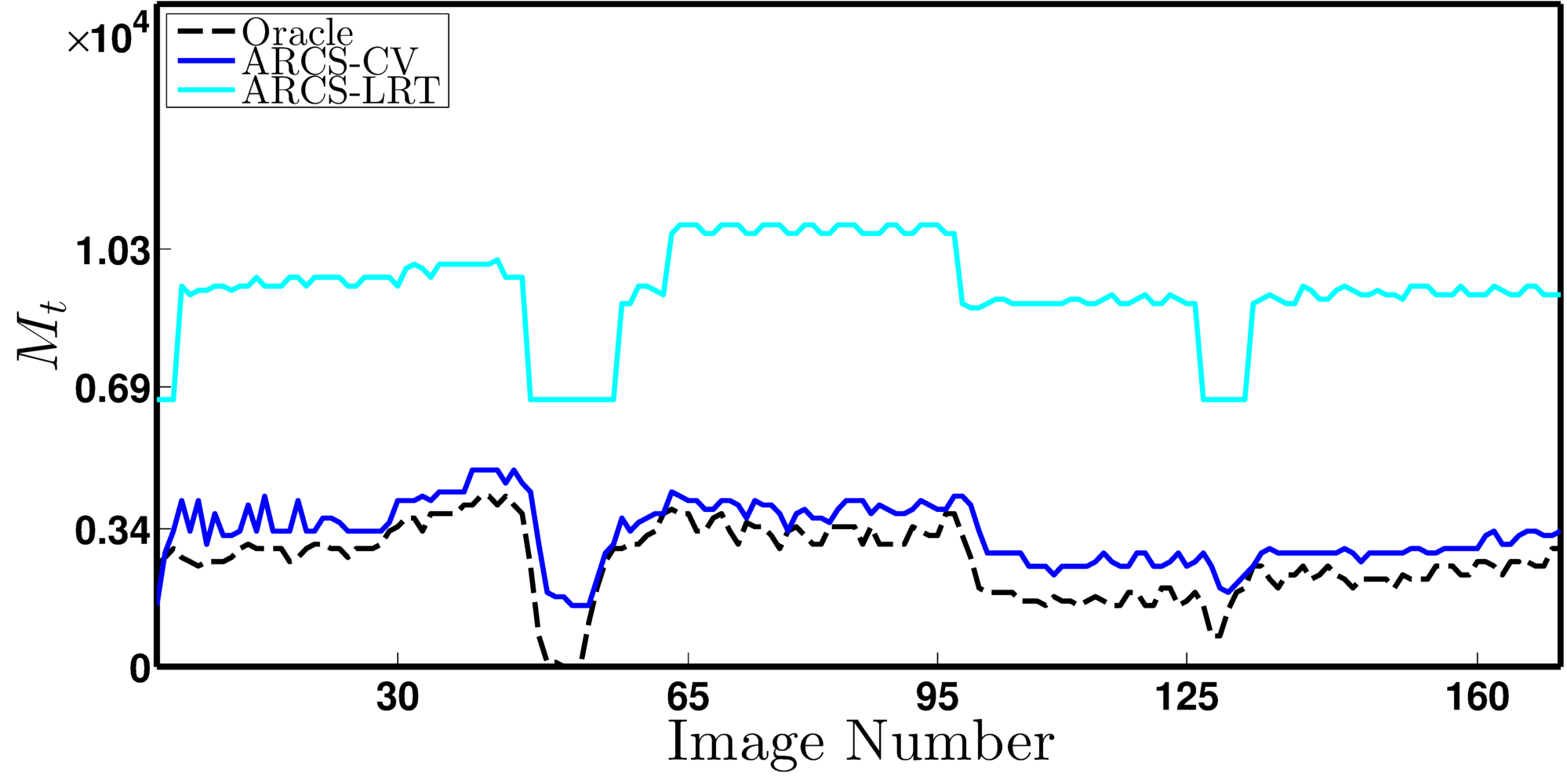} \label{fig::experiments_results_PETS2009_M}}
  	\subfloat[]{\includegraphics[width=0.32\textwidth]{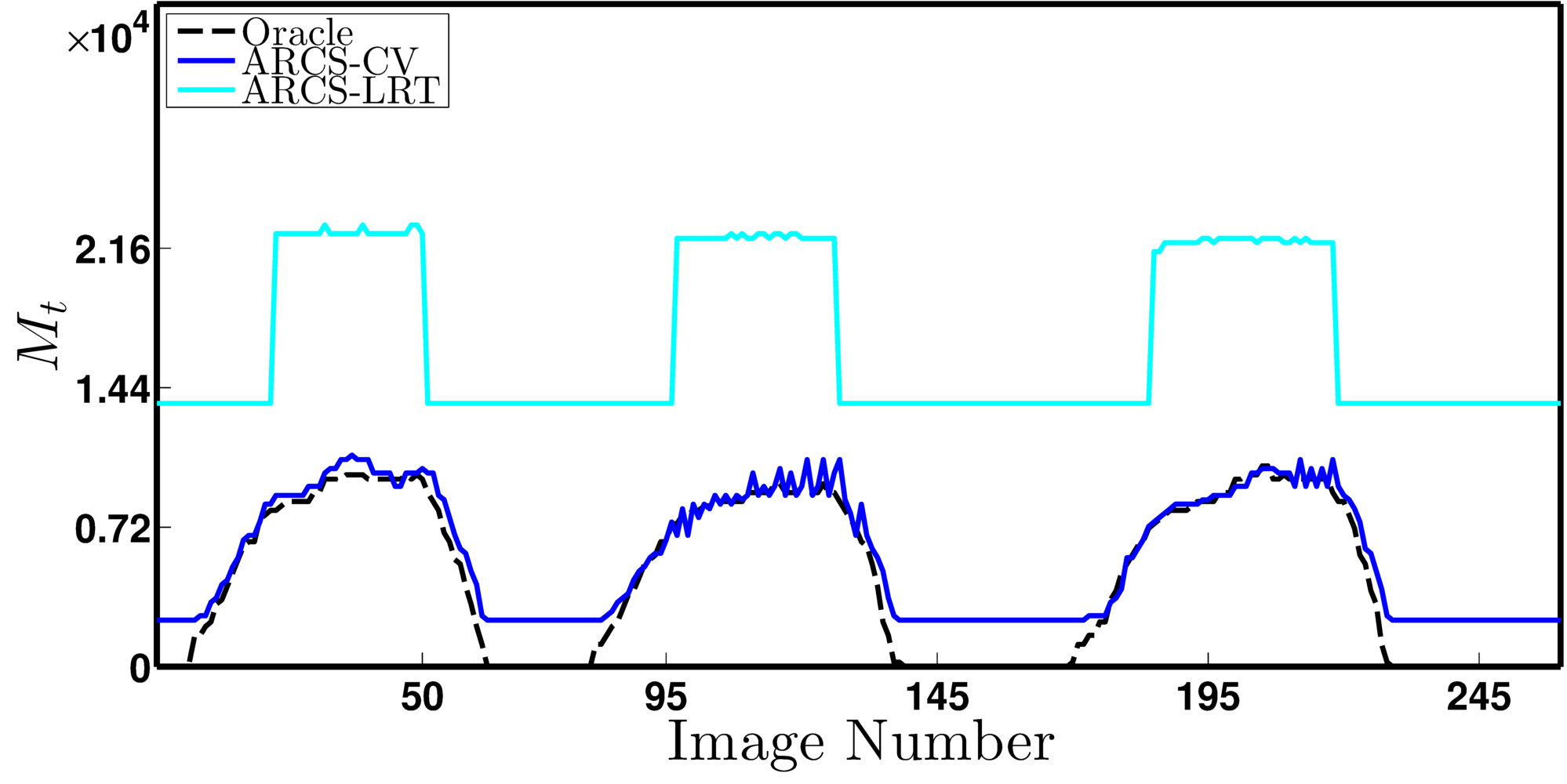} \label{fig::experiments_results_convoy2_M}}
  	  	
  	\vspace{10pt}
  	
  	\subfloat[]{\includegraphics[width=0.32\textwidth]{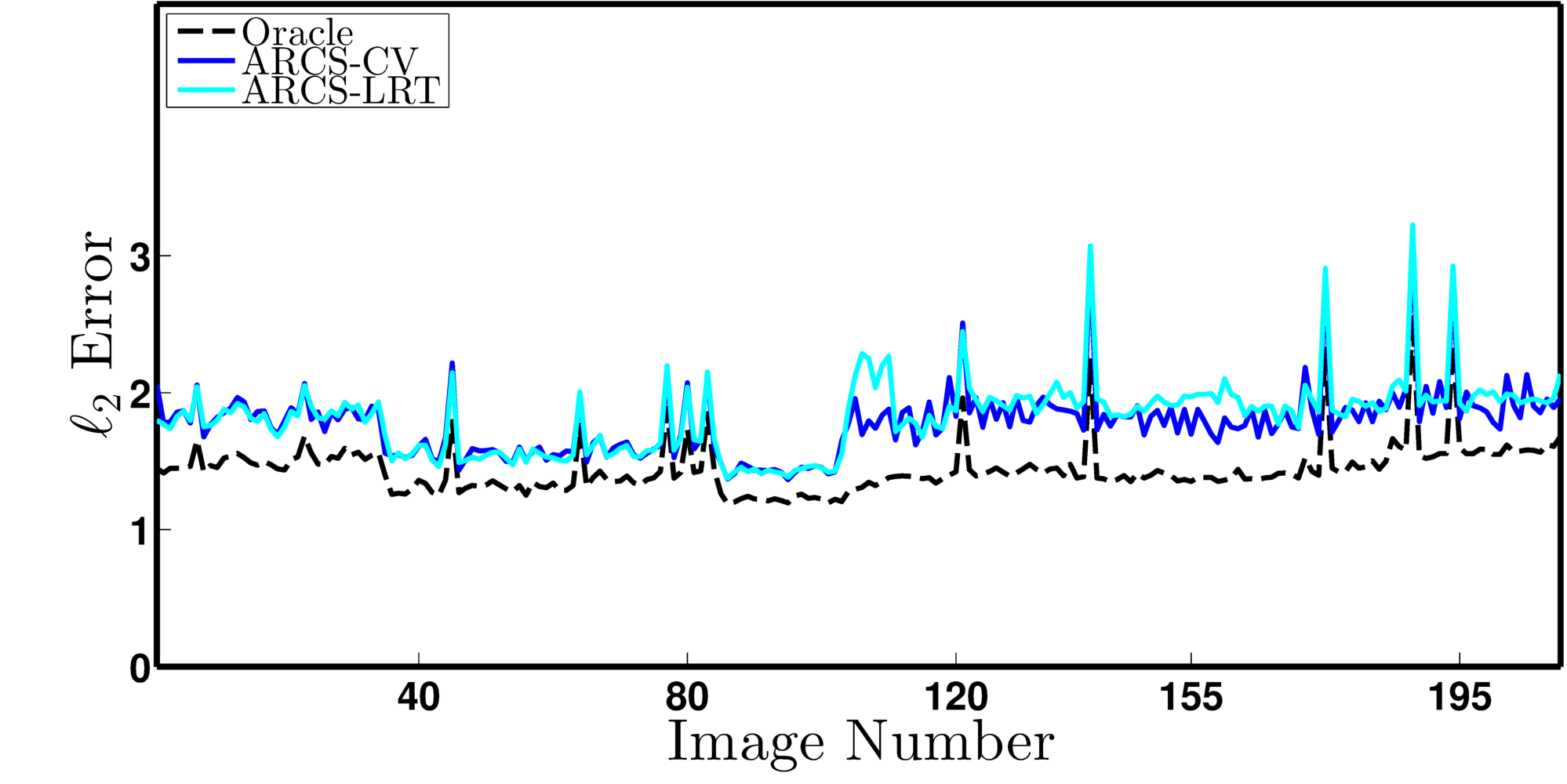} \label{fig::experiments_results_markercam_e}}
  	\subfloat[]{\includegraphics[width=0.32\textwidth]{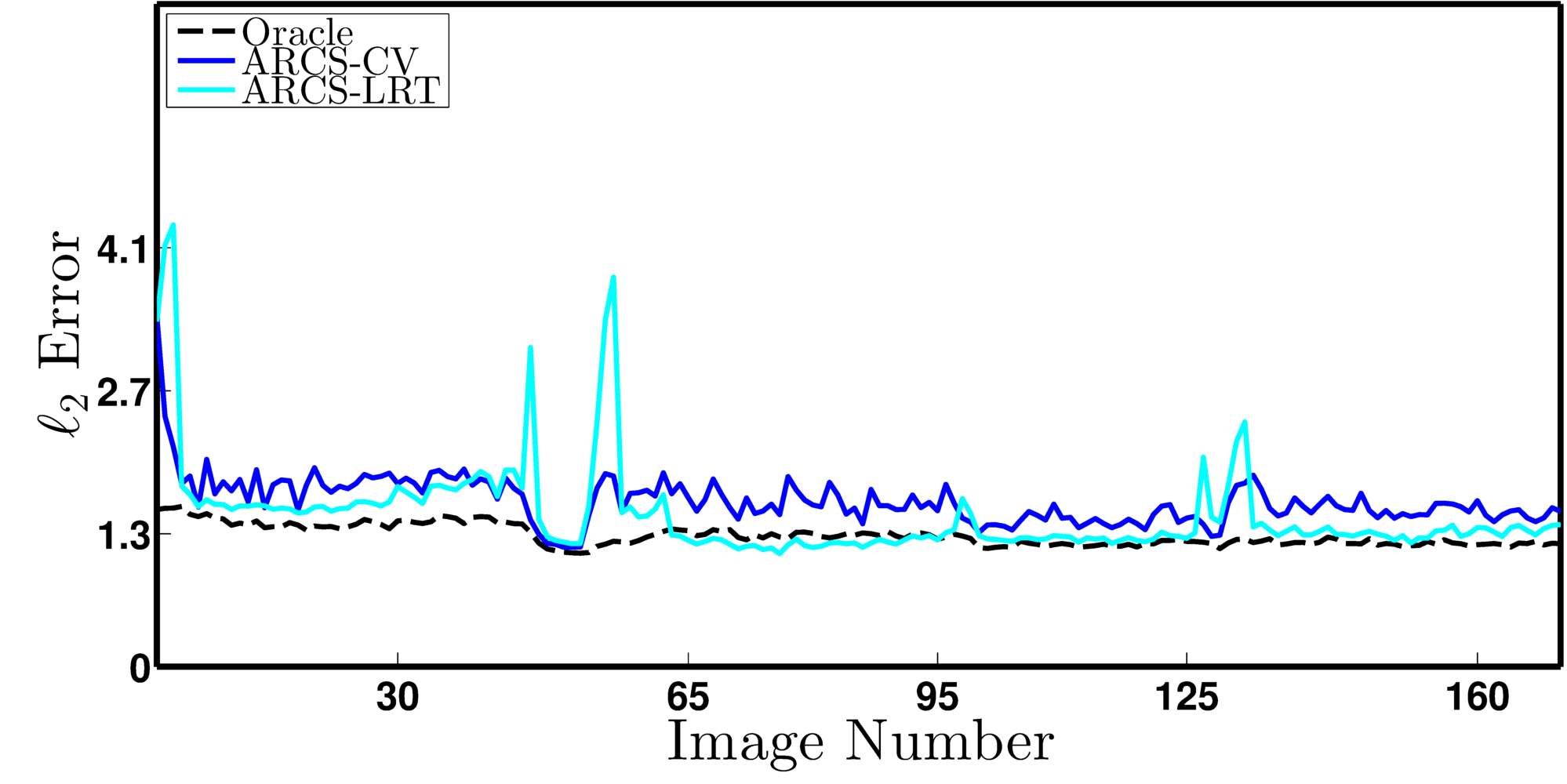} \label{fig::experiments_results_PETS2009_e}}
  	\subfloat[]{\includegraphics[width=0.32\textwidth]{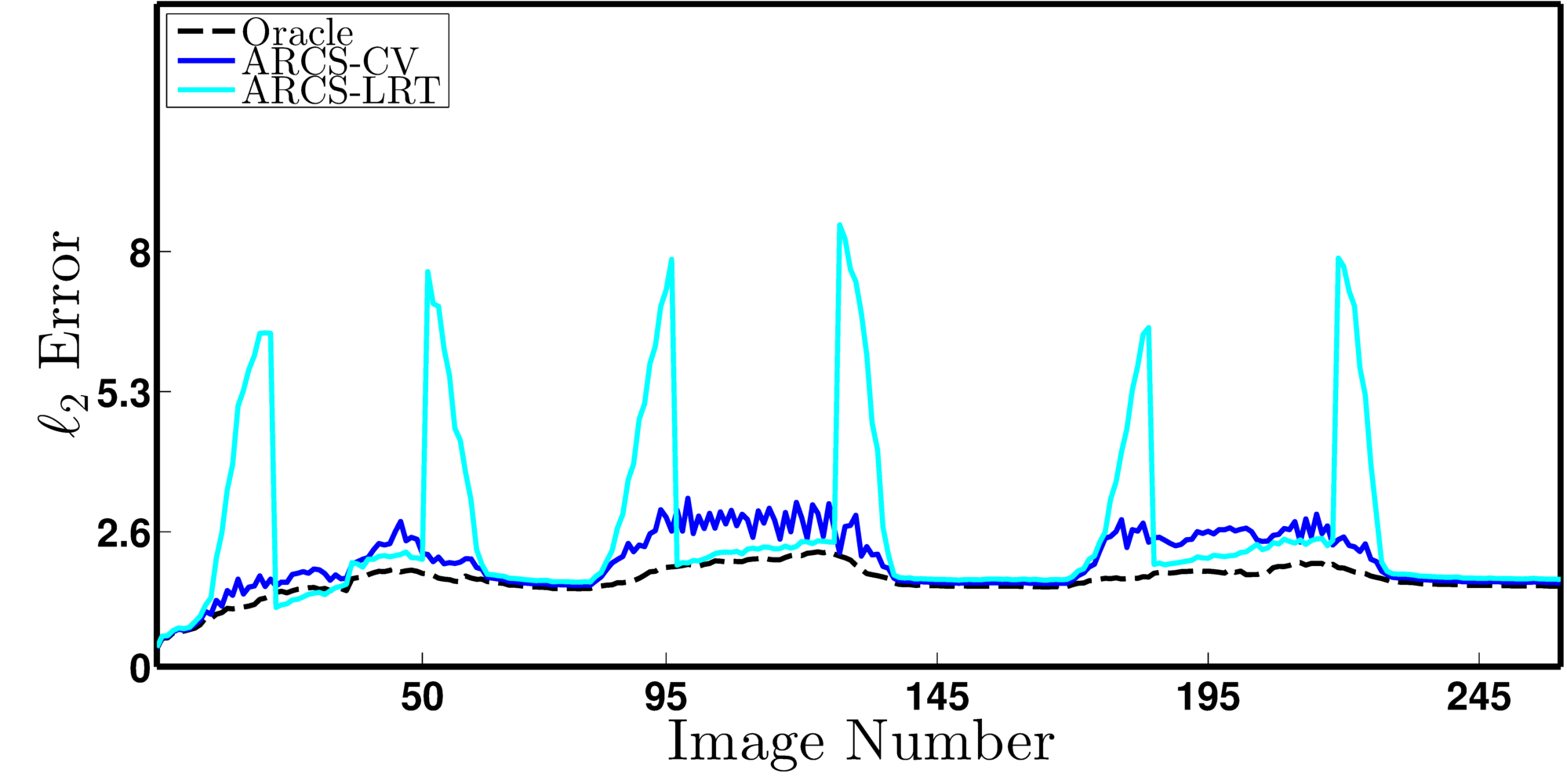} \label{fig::experiments_results_convoy2_e}}
  	  	
	\caption{\footnotesize{Performance of adaptive CS strategies for the \texttt{marker\_cam} (column one), \texttt{PETS2009\_S2L1} (column two), and \texttt{convoy2} (column three) video sequences. In the first row, $\hat{s}_t$ is used to denote the sparsity estimate used by each strategy.  In row two, $M_t$ is used to denote the total number of measurements that must be acquired.  The $\ell_2$ reconstruction error is plotted in row three.}}
	\label{fig::experiments_results}
\end{figure*}

In order to provide some context in which to interpret the results from our ARCS methods, we present them alongside those from the best-case sensing strategy: \emph{oracle CS}.  Oracle CS uses the true value of $s_t$ as its sparsity estimate, which is impossible to obtain in practice.  We compare the average measurement rates and foreground reconstruction errors for the three methods (oracle, ARCS-CV, and ARCS-LRT) in Table \ref{table::experiments_strategycompare}, and show the more detailed dynamic behavior in Figure \ref{fig::experiments_results}.   Note that the measurement values reported for the ARCS algorithms include the necessary overhead for the side information (i.e., the cross-validation and low-resolution measurements).

\begin{table*}
	\centering
	\caption{Experimental comparison of adaptive compressive sensing measurement strategies (oracle, ARCS-CV, ARCS-LRT)}
	\begin{tabular}{ |l!{\vrule width 1.5pt}c|c|c!{\vrule width 1.5pt}c|c|c| }
		\hline
 		& \multicolumn{3}{c!{\vrule width 1.5pt}}{Average \# of Measurements ($\bar{M}/N^2$)} & \multicolumn{3}{c|}{Average Reconstruction Error ($\ell_2$)} \\
		\cline{2-7}
		& Oracle & ARCS-CV & ARCS-LRT & Oracle & ARCS-CV & ARCS-LRT \\
		\noalign{\hrule height 1.5pt}
		\texttt{marker\_cam} & 0.0598 & 0.0939 & 0.3356 & 1.4388 & 1.7802 & 1.8229 \\
		\hline
		\texttt{PETS2009\_S2L1} & 0.1209 & 0.1530 & 0.4238 & 1.2811 & 1.6181 & 1.4911 \\
		\hline
		\texttt{convoy2} & 0.0997 & 0.1251 & 0.3627 & 1.6573 & 2.0296 & 2.6137 \\
		\hline
	\end{tabular}
	\label{table::experiments_strategycompare}
\end{table*}

We first observe that the ARCS-LRT algorithm uses a significantly larger measurement rate than any of the others.  This is due to the necessary overhead for the low-resolution side information.  In our experiments, we used $L=N/2$, i.e. $M_t$ is at least $25\%$ of $N^2$.  A smaller $L$ could be selected at the risk of poorer low-resolution tracking.  The ARCS-CV algorithm performs much better in terms of measurement rate since the side-information overhead is relatively small (for all datasets, $r$ is less than $2\%$ of $N^2$).

It can also be seen that the ARCS-LRT sparsity estimate lags behind the true foreground sparsity for those images in which an object is entering or exiting the camera's field-of-view but not fully visible.  The phenomenon is especially visible in the third column (\texttt{convoy2}) of Figure \ref{fig::experiments_results}.  It is due to the fact that we have manually imposed the condition that the object cannot be tracked unless it is fully visible.  This leads to the large spikes in foreground reconstruction error.  However, when the object becomes fully visible, the low-resolution tracks provide the algorithm with enough information to monitor the high-resolution signal sparsity and the effect disappears.

\subsection{Steady-State Behavior}
\begin{figure}
	\centering
	\subfloat[]{\includegraphics[width=0.45\textwidth]{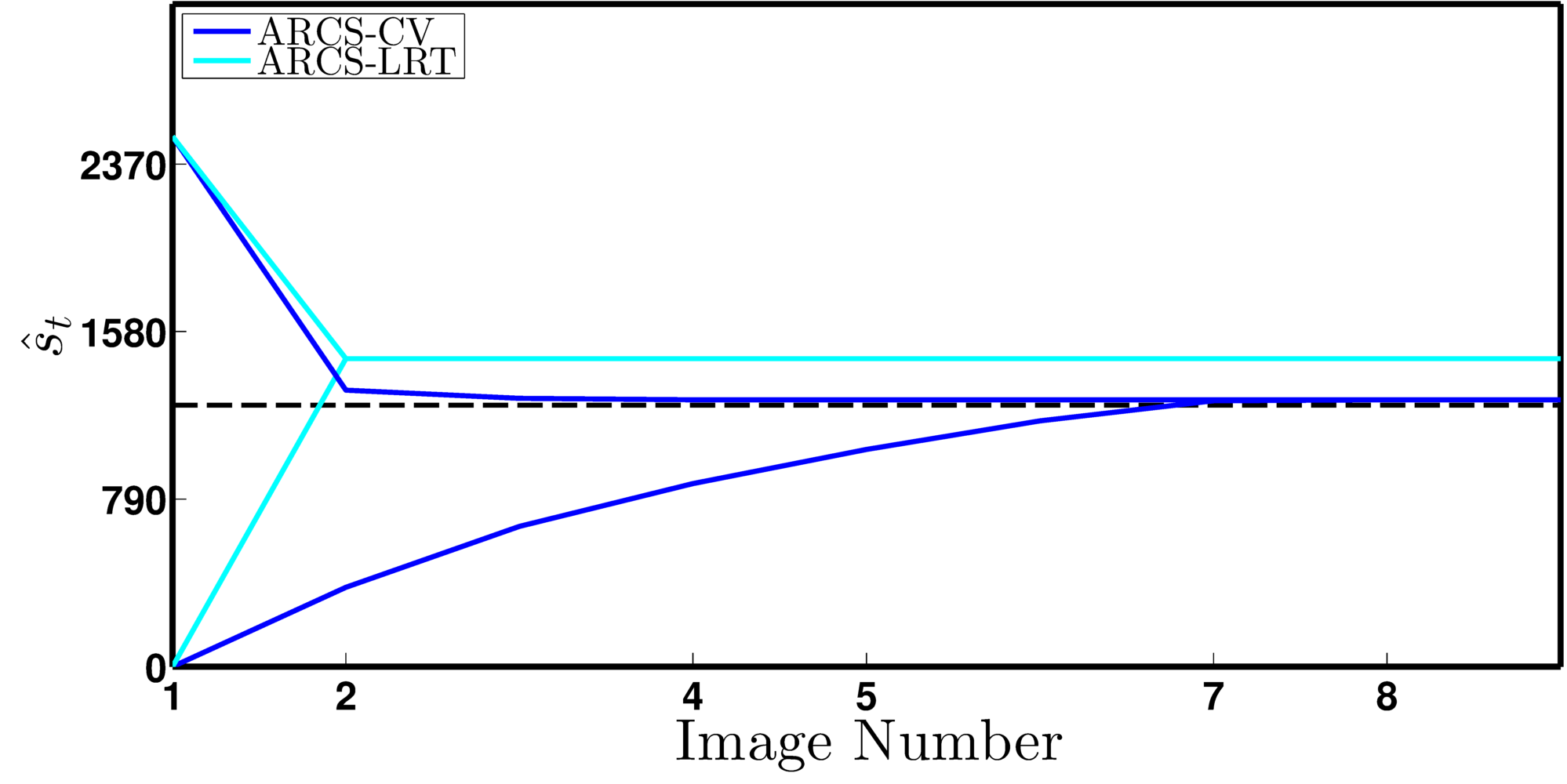} \label{fig::experiments_steady_convoy2_s}}
	
	\subfloat[]{\includegraphics[width=0.45\textwidth]{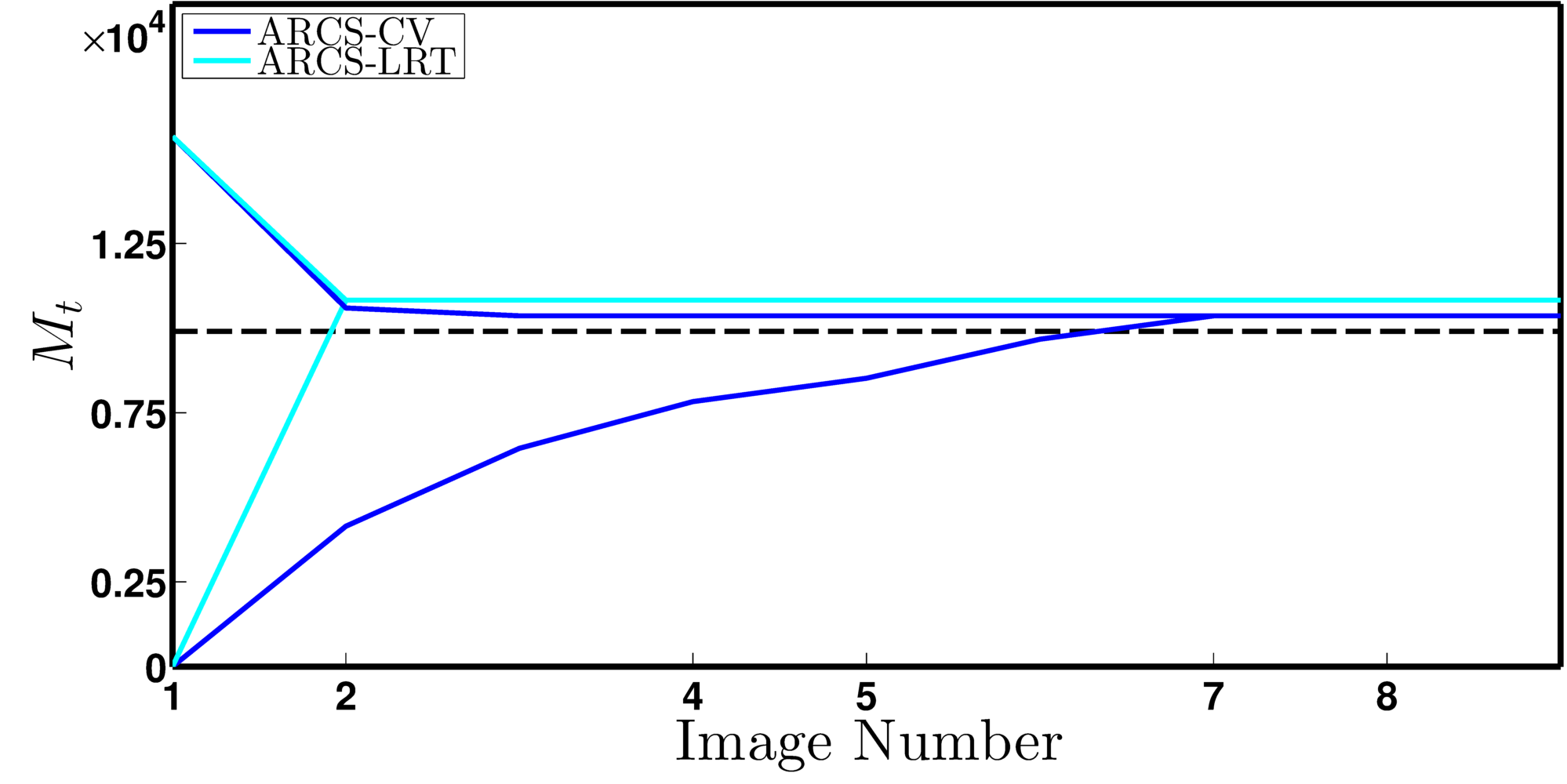} \label{fig::experiments_steady_convoy2_M}}
	
	\subfloat[]{\includegraphics[width=0.45\textwidth]{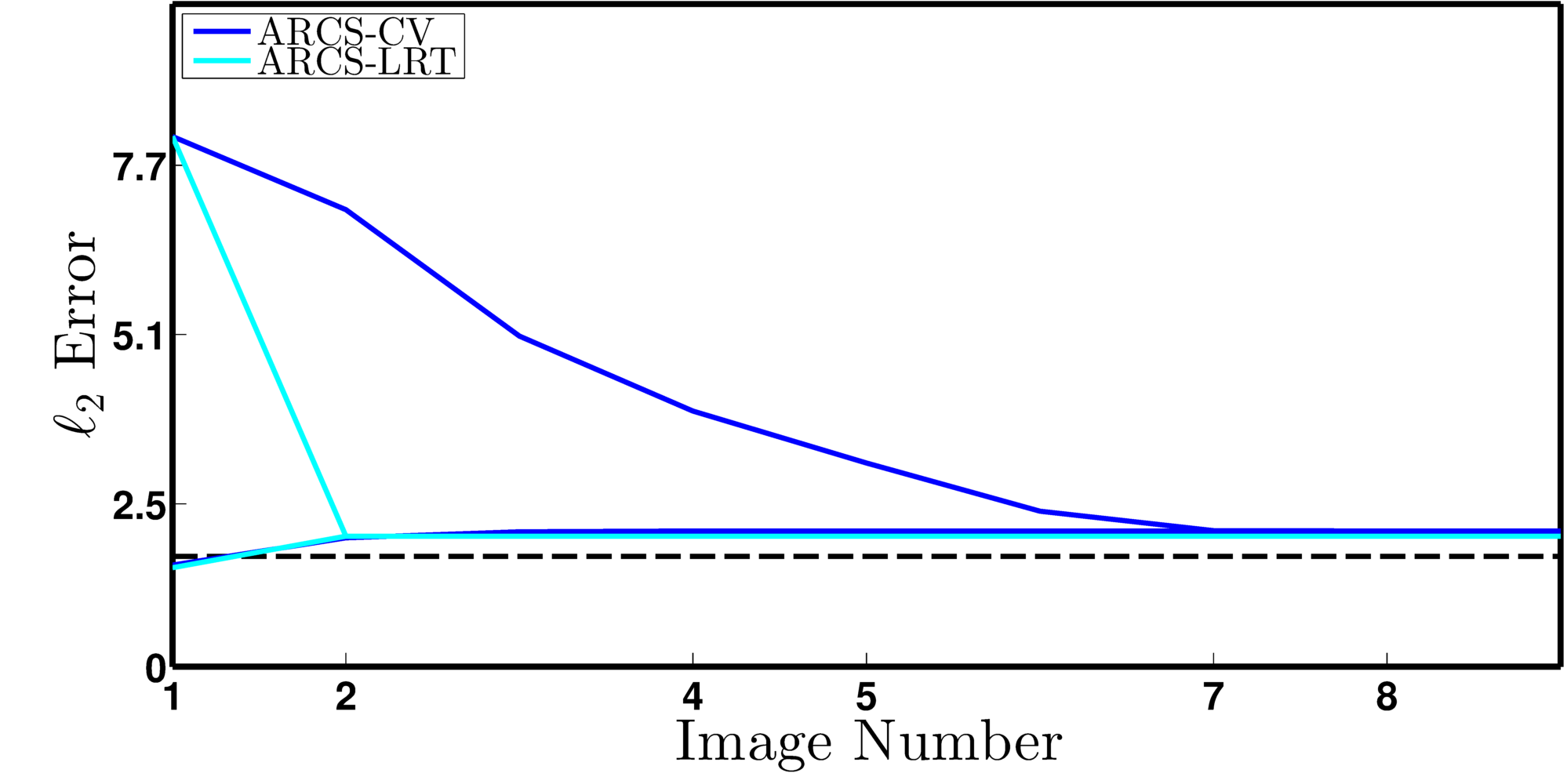} \label{fig::experiments_steady_convoy2_e}}
		
	\caption{\footnotesize{Steady-state behavior for both ARCS algorithms using a video sequence constructed by repeating a single image selected from the \texttt{convoy2} dataset.  For each algorithm, two experimental paths are shown: one generated by initializing the sparsity estimate such that it is too small ($s_1 << s$), and the other generated by initializing the sparsity estimate such that it is too large ($s_1 >> s$).}}
	\label{fig::experiments_steady}
\end{figure}

We analyzed the behavior of our ARCS methods when the signal under observation is static (i.e., $\mathbf{f}_t = \mathbf{f}$ for all $t$).  To do so, we created a synthetic data sequence by repeating a single image in the \texttt{convoy2} data set for which $s = 1233$.  Figure \ref{fig::experiments_steady} shows the behavior of each algorithm when the initial sparsity estimate, $\hat{s}_1$, is wrong.  For each method, we ran two experiments.  For the first one, we initialized the sparsity estimate using a value that was too low ($\hat{s}_1 = 0$).  For the second one, we initialized with a value that was too high ($\hat{s}_1 = 2500$).  Note that both methods are able to successfully adapt to the true value of $s$, and the ARCS-LRT method adapts very quickly (requiring only a single image) due to the immediate availability of the low-resolution track.

\subsection{ARCS-LRT and Automatic Tracking}
\begin{figure}
	\centering
	\subfloat[]{\includegraphics[width=0.45\textwidth]{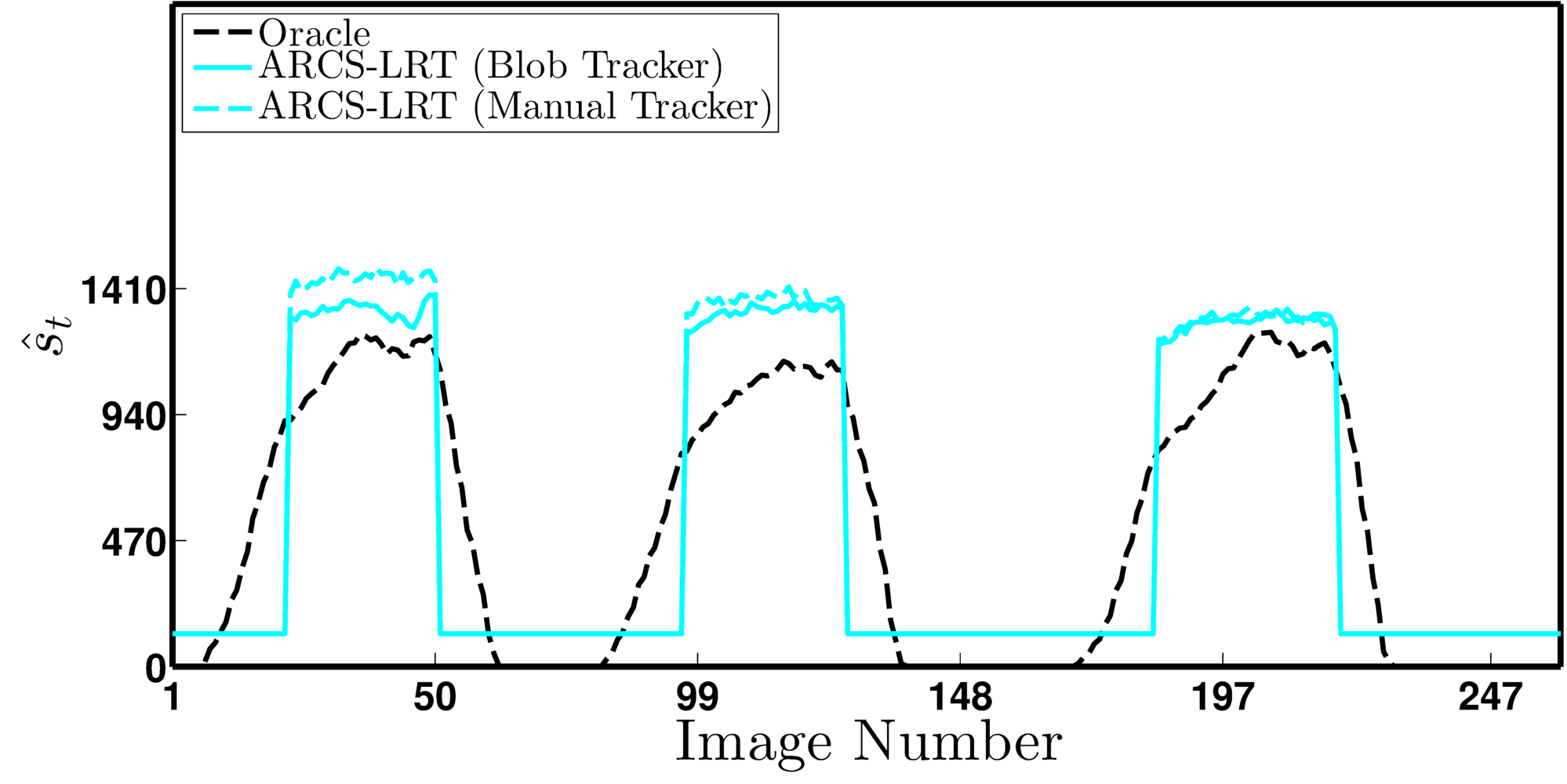} \label{fig::experiments_trackeffect_convoy2_s}}
	
	\subfloat[]{\includegraphics[width=0.45\textwidth]{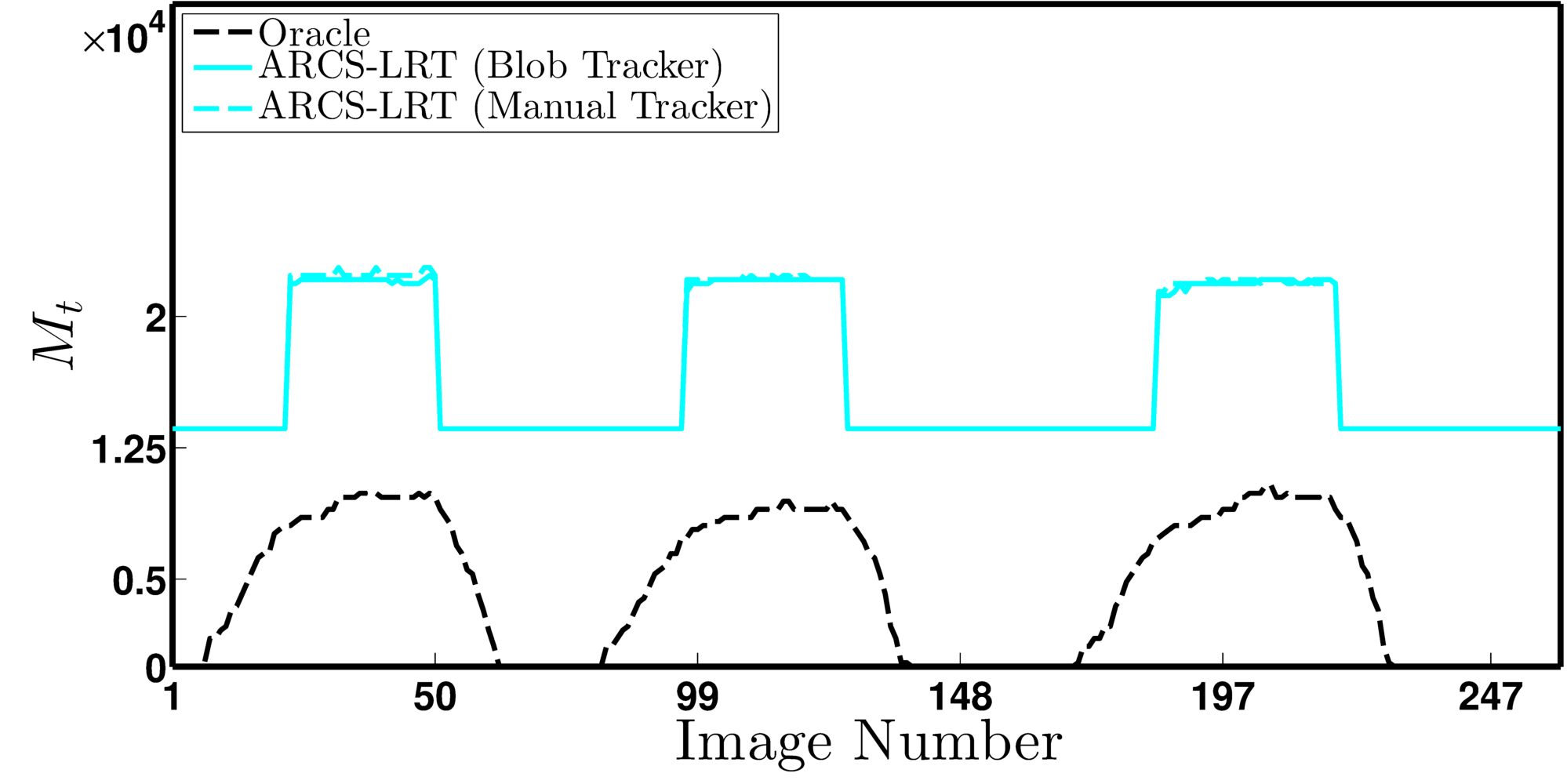} \label{fig::experiments_trackeffect_convoy2_M}}
	
	\subfloat[]{\includegraphics[width=0.45\textwidth]{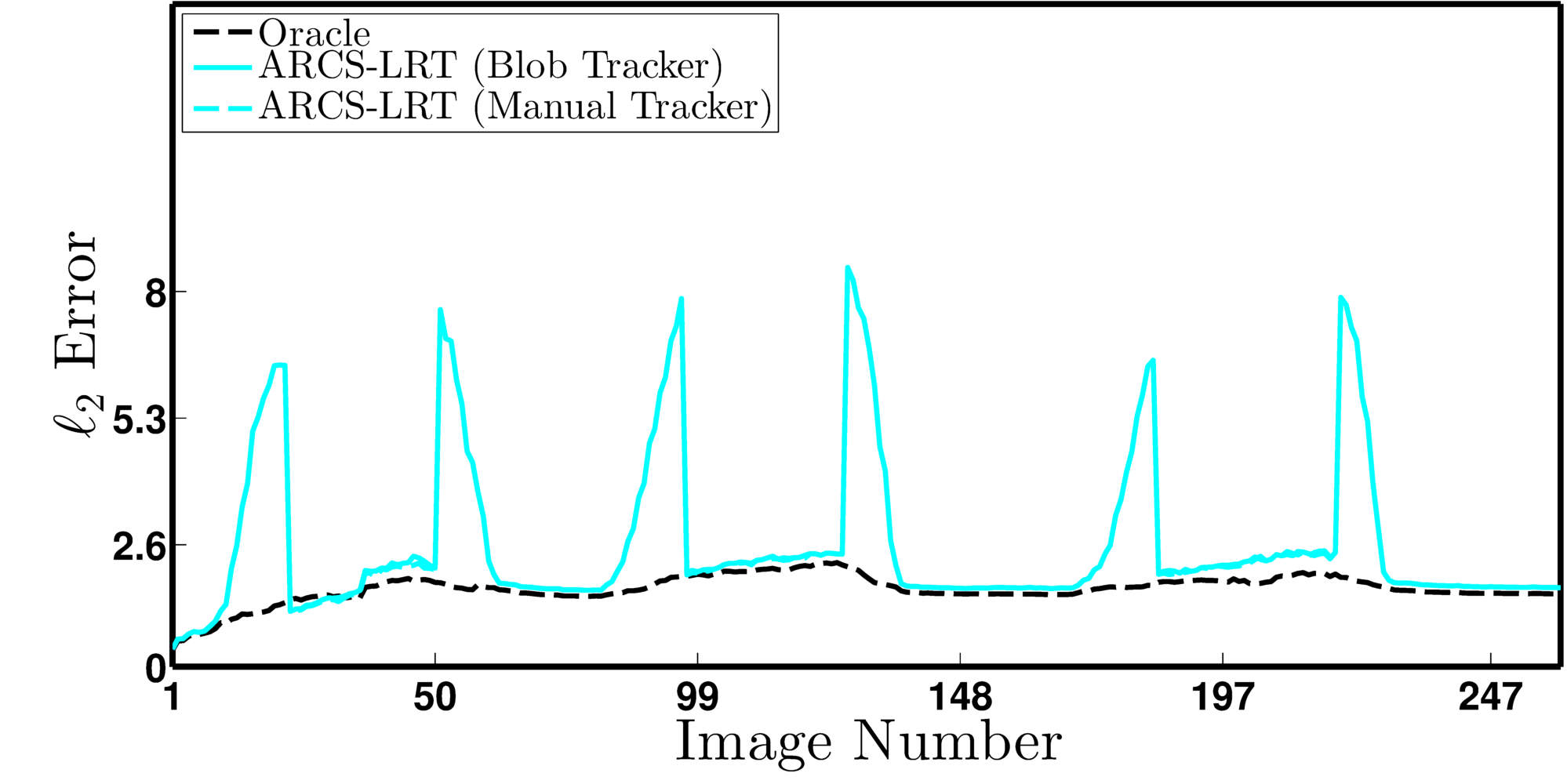} \label{fig::experiments_trackeffect_convoy2_e}}
	
	\caption{\footnotesize{Effect of manual vs. automatic blob tracking on the behavior of the ARCS-LRT method for the \texttt{convoy2} dataset.}}
	\label{fig::experiments_trackeffect}
\end{figure}

We also investigated the effect of using low-resolution tracks obtained via an automatic method.  To do so, we implemented a simple blob tracker in MATLAB for the \texttt{convoy2} sequence and used the resulting tracks in the ARCS-LRT framework.  A comparison of algorithm performance between using automatic tracks and our manually-marked tracks is shown in Figure \ref{fig::experiments_trackeffect}.  Given the negligible effect of the blob tracker on the behavior of ARCS-LRT, we would not expect more sophisticated automatic tracking techniques to negatively affect performance.

\section{Summary and Future Work}
\label{secn::summary}
We have described two techniques for using side information to adjust the measurement rate of a dynamic compressive sensing system.  These techniques were developed in the specific context of using this system for video background subtraction.  The first technique involves collecting side information in the form of a small number of extra cross-validation measurements and using an error bound to infer underlying signal sparsity.  The second method uses side information from a secondary, low-resolution, traditional camera in order to infer the sparsity of the high-resolution images.  In either case, we used a pre-computed phase diagram as a lookup table to map sparsity estimates to minimal compressive measurement rates.  We validated these techniques on real video sequences using practical approximations for theoretical quantities.

This work provides a framework that allows for numerous extensions:
\begin{itemize}
\item It may be possible to achieve more optimal measurement rates by modifying the decoder.  For example, using techniques like those developed by Vaswani \emph{et al.} \cite{Vaswani2010a}, the phase diagrams we use could be updated.

\item In addition to modifying the number of rows, the \emph{content} of the measurement matrix could be adjusted between acquisitions as well.  Such a strategy would be theoretically similar to the previously-discussed work of Duarte-Carvajalino \cite{Duarte-Carvajalino2012} \emph{et al.} and others \cite{Averbuch2012} \cite{Ji2008} \cite{Chou2009} \cite{Haupt2012}, but with a fixed measurement budget at each time instant that would change from acquisition to acquisition.

\item The assumption that the side sensor in ARCS-LRT is co-located with the compressive camera could be removed.  This might involve a more complicated mapping function (\ref{eqn::method2_tracktosparsity}) that also incorporates knowledge of the geometrical relationship between the two sensors.
\end{itemize}
\section*{Acknowledgements}
The authors would like the thank Vishal Patel, Rachel Ward, Mark Davenport, and Aswin Sankaranarayanan for their correspondence during the development of this paper.

\bibliographystyle{IEEEtran}
\bibliography{IEEEabrv,CSSI_TIP}


\end{document}